\documentclass[runningheads]{llncs}

\usepackage[mobile]{eccv}

\usepackage{eccvabbrv}

\usepackage{graphicx}
\usepackage{booktabs}

\usepackage[accsupp]{axessibility}  

\usepackage{multirow}
\usepackage{makecell}
\usepackage[table]{xcolor}
\usepackage{caption}
\usepackage{float}
\usepackage{tablefootnote}
\usepackage{threeparttable}

\usepackage[dvipsnames]{xcolor} 
\usepackage{pifont,xcolor}
\usepackage[final]{microtype}
\usepackage{tcolorbox}
\tcbuselibrary{breakable}

\definecolor{SoftRed}{RGB}{220,50,70}
\definecolor{DeepTeal}{RGB}{0,102,150}
\definecolor{Amber}{RGB}{255,140,0}
\definecolor{Forest}{RGB}{0,150,90}

\newtcolorbox[auto counter]{promptbox}[2][]{
    colback=gray!5,
    colframe=gray!50,
    fonttitle=\bfseries,
    title={Prompt \thetcbcounter: #2},
    breakable,
    arc=3pt,
    boxrule=0.5pt,
    left=6pt, right=6pt, top=6pt, bottom=6pt,
    #1
}

\newif\ifshowcomments

\showcommentstrue 

\ifshowcomments
  \newcommand{\KGnote}[1]{{\color{magenta}#1}}
  
  \newcommand{\JBnote}[1]{{\color{DeepTeal}#1}}
  
\else
  \newcommand{\KGnote}[1]{}
  
  \newcommand{\JBnote}[1]{}
  
\fi

\usepackage{hyperref}
\usepackage{orcidlink}

\begin{document}


\title{UniversalVTG: A Universal and Lightweight Foundation Model for Video Temporal Grounding} 

\titlerunning{UniversalVTG}

\author{Joungbin An, Agrim Jain, Kristen Grauman}
\authorrunning{J. An et al.}
\institute{The University of Texas at Austin}

\maketitle
\vspace{-3mm}
\begin{center}
    \includegraphics[width=0.75\linewidth]{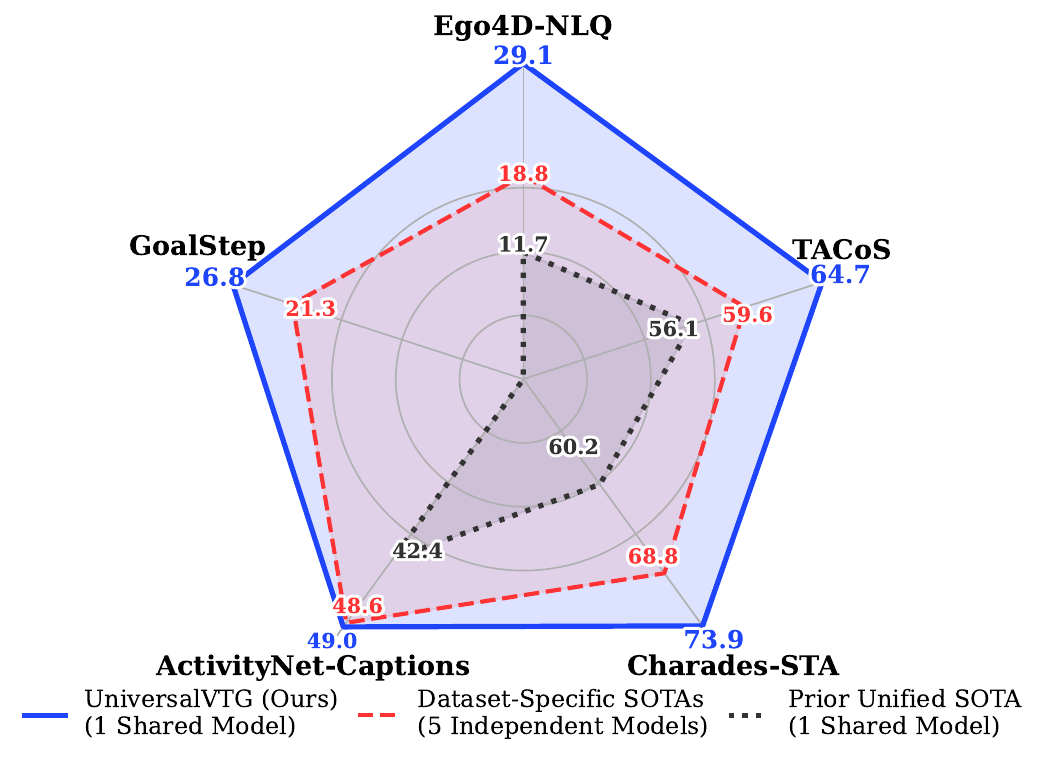}
    \vspace{-3mm}
    \captionof{figure}{
    Universal without compromise: a single \textsc{UniversalVTG} checkpoint rivals dataset-specific SOTA.
    }
    \label{fig:teaser}
    \vspace{-3mm}
\end{center}

\begin{abstract}
Video temporal grounding (VTG) is typically tackled with dataset-specific models that transfer poorly across domains and query styles. Recent efforts to overcome this limitation have adapted large multimodal language models (MLLMs) to VTG, but their high compute cost and limited video context still hinder long-video grounding. We instead scale \emph{unified supervision} while keeping the model lightweight. We present \textbf{UniversalVTG}, a single VTG model trained with large-scale cross-dataset pretraining. An offline Query Unifier canonicalizes heterogeneous query formats into a shared declarative space, reducing linguistic mismatch and preventing the negative transfer observed under naïve joint training. Combined with an efficient grounding head, UniversalVTG scales to long, untrimmed videos. Across diverse benchmarks---GoalStep-StepGrounding, Ego4D-NLQ, TACoS, Charades-STA, and ActivityNet-Captions---one UniversalVTG checkpoint achieves state-of-the-art performance versus dedicated VTG models. Moreover, despite being $>100\times$ smaller than recent MLLM-based approaches, UniversalVTG matches or exceeds their accuracy on multiple benchmarks, offering a practical alternative to parameter-heavy MLLMs.\footnote{Project webpage: \url{https://vision.cs.utexas.edu/projects/universalvtg}.}
\vspace{-1mm}
\end{abstract}

\section{Introduction}
\label{sec:intro}

The ability to temporally localize open-language descriptions in untrimmed video, known as Video Temporal Grounding (VTG), underpins emerging applications in video search, summarization, and human-robot interaction. Whether a user is retrieving \emph{``the moment I left the stove on''} from egocentric footage, or a creator is searching for \emph{``the decisive goal,''} the system must align free-form text with precise temporal boundaries. While VTG has progressed from early models designed for short clips~\cite{anne2017localizing,soldan2021vlg, gao2017tall,zhang2020learning,mun2020LGI,zhang2020span,lei2021detecting} to architectures handling hour-scale videos~\cite{hou2022cone,snag,lu2025decafnet,an2025hieramamba}, current solutions remain difficult to deploy as general-purpose systems in open-world settings.

A key practical limitation is that VTG models are typically engineered around individual benchmarks~\cite{anne2017localizing,soldan2021vlg, gao2017tall,zhang2020learning,mun2020LGI,zhang2020span,lei2021detecting,hou2022cone,snag,lu2025decafnet,an2025hieramamba,hannan2024rgnet}. Differences in domain, temporal granularity, and query style routinely force dataset-specific design choices. For example, Charades-STA features short, third-person action clauses (e.g., \emph{``person runs to the window''}), whereas Ego4D-NLQ pairs long egocentric videos with conversational, state-driven questions (e.g., \emph{``Where was the green cloth before I picked it?''}). Because benchmark-centric supervision encourages models to overfit to these dataset-specific linguistic conventions and visual regimes, a model optimized for one dataset typically degrades substantially when applied to another (Table~\ref{tab:cross_dataset}). In effect, the community has converged on an $N$-datasets to $N$-models paradigm, preventing the deployment of a single system capable of handling heterogeneous video sources without manual reconfiguration.

A recent response to these limitations is adapting large multimodal language models (MLLMs)~\cite{wang2024qwen2} for VTG~\cite{ren2024timechat, wang2024hawkeye, pramanick2025enrich, li2025universal}. While promising, state-of-the-art MLLMs are parameter-heavy and operate with restricted visual contexts. To process long videos, typical pipelines must chunk the video and repeatedly re-encode frames~\cite{ren2024timechat, li2025universal, wang2024hawkeye}, which is computationally prohibitive and scales poorly. In most practical settings, processing thousands of dense video tokens through a multi-billion-parameter model, let alone repeatedly invoking it across sliding windows, is computationally prohibitive and fundamentally misaligned with the demands of continuous real-world deployment.

Instead of scaling parameter count to overcome domain gaps, we advocate scaling \emph{unified supervision} for highly efficient, lightweight architectures. To this end, we present \textbf{UniversalVTG}, a universal and lightweight model for video temporal grounding. We enable a single architecture to handle all domains by processing diverse video inputs through a shared visual-temporal backbone. Crucially, however, we resolve the negative transfer that traditionally plagues cross-dataset training by introducing a \emph{Query Unifier} that canonicalizes heterogeneous dataset queries into a standardized semantic space. This semantic harmonization is the key to unlocking generalization, allowing the model to learn synergistically from heterogeneous datasets so that diverse domains positively reinforce one another. We then scale this unification by aggregating over one million query--segment pairs for pretraining. By training on this massive, harmonized corpus, our compact model learns intent-level grounding invariant to stylistic shifts, utilizing a scale of VTG-specific supervision that prior lightweight models have not exploited.

Empirically, UniversalVTG establishes strong performance with a single set of weights across five major benchmarks spanning first- and third-person perspectives, as well as short and long videos: GoalStep~\cite{song2023ego4d}, Ego4D-NLQ~\cite{grauman2022ego4d}, TACoS~\cite{regneri2013grounding}, Charades-STA~\cite{sigurdsson2016hollywood}, and ActivityNet-Captions~\cite{krishna2017dense}. Notably, despite being over two orders of magnitude smaller than recent MLLM-based VTG approaches~\cite{ren2024timechat, zeng2024timesuite,guo2025vtg,chen2024timemarker,zhang2025videollama,pramanick2025enrich,li2025universal}, UniversalVTG achieves comparable or superior results. In summary, our contributions are:
\begin{itemize}
    \item \textbf{UniversalVTG}, a single lightweight model trained jointly to generalize across heterogeneous domains, spanning both short and long videos without dataset-specific tuning;
    \item a \textbf{unified training framework} utilizing a Query Unifier to canonicalize text inputs, effectively resolving language-style mismatches and preventing negative transfer during joint training;
    \item \textbf{large-scale harmonized pretraining:} we show that structurally standardizing the instruction space across \textgreater1M query--segment pairs is the key to unlocking cross-domain generalization for lightweight architectures, establishing a scalable alternative to MLLMs;
    \item experiments demonstrating \textbf{strong generalization and state-of-the-art performance} across multiple VTG benchmarks with one universal model.
\end{itemize}

\section{Related Work}

\noindent\textbf{Video Temporal Grounding.}
Video temporal grounding (VTG) localizes the start and end timestamps of a natural-language query in untrimmed video, enabling applications in episodic memory, video editing, and human--robot interaction.
Much of the early literature targets \emph{short} clips, where the search space is modest and evidence is dense, using either proposal ranking~\cite{anne2017localizing,gao2017tall,zhang2020learning,yuan2019semantic,wang2021structured,soldan2021vlg,chen2023joint} or direct boundary regression~\cite{ghosh2019excl,zeng2020dense,mun2020LGI,zhang2020span}.
Recent short-video VTG is dominated by DETR-like set prediction formulations~\cite{carion2020end,lei2021detecting,moon2023query,moon2023correlation,gordeev2024saliency}, exemplified by Moment-DETR~\cite{lei2021detecting}, with follow-up work improving alignment via stronger priors and attention mechanisms~\cite{jang2023knowing,moon2023query,moon2023correlation}.

Long videos introduce a qualitatively different regime: relevant evidence is sparse and can be separated by minutes, creating a ``needle-in-a-haystack'' challenge.
Early \emph{long-video temporal grounding} (LVTG) systems reduce computation through truncation or fixed-length pooling~\cite{zhang2020learning,zhang2020span,soldan2021vlg,lei2021detecting,ramakrishnan2023spotem}, often discarding fine temporal detail.
Subsequent methods preserve more context via coarse-to-fine pipelines or fixed-size sliding windows~\cite{hou2022cone,hannan2024rgnet,pan2023scanning}, though window boundaries can disrupt temporal coherence.
More recent approaches introduce multi-scale modeling and windowed attention~\cite{zhang2022actionformer,snag,lu2025decafnet,feng2025OSGNet}, while HieraMamba~\cite{an2025hieramamba} improves scalability further using state-space models with hierarchical token compression to achieve effective linear-time grounding on long untrimmed videos. 

Building a universal model requires overcoming both temporal and linguistic divides across heterogeneous datasets. Early consolidation efforts like UniVTG~\cite{lin2023univtg} aggregate diverse \emph{tasks}, but still rely on task-specific decoding heads and ignore domain-level language shifts, rendering them insufficiently unified. Furthermore, while recent lightweight architectures~\cite{snag, lu2025decafnet, an2025hieramamba} can scale temporally within isolated domains, naively merging datasets introduces conflicting query conventions that induce negative transfer. We address this dual challenge by pairing an efficient linear-time backbone with a semantic Query Unifier at the supervision interface. This structural harmonization enables a single, shared-weight checkpoint to generalize seamlessly across first-person, third-person, short, and long videos.

\vspace{2mm}
\noindent\textbf{Temporal Grounding with Multi-modal Language Models.}
The rise of multi-modal large language models (MLLMs) has motivated their use for temporal grounding, exploiting strong language understanding and broad visual--semantic priors~\cite{wang2024qwen2, bai2025qwen3,clark2026molmo2}.
A key difference among MLLM-based VTG methods is how they access temporal information.
Some approaches are \emph{time-agnostic} or weakly time-aware, regressing normalized boundaries from a fixed number of frames~\cite{huang2024vtimellm}, sometimes via special temporal tokens~\cite{huang2024lita}.
Others encode time \emph{implicitly} via timestamp-aware embeddings or positional schemes~\cite{ren2024timechat,zeng2024timesuite,guo2025vtg,bai2025qwen3}.
A third family uses \emph{explicit temporal marking}, attaching textual timestamps to frames so grounding becomes retrieval over tagged tokens~\cite{meinardus2024surprising,chen2024timemarker,zhang2025videollama}.
Recent work further couples MLLMs with lightweight boundary heads, e.g., enriching queries with video context to decode boundaries from a learned interval token~\cite{pramanick2025enrich}. Most recently, UniTime~\cite{li2025universal} targets long videos using a coarse-to-fine sliding-window pipeline to propose and refine candidate regions, further improving performance with large-scale video--language training.

While MLLMs offer flexibility, their long-video deployment is computationally prohibitive due to massive visual token counts and expensive multi-pass inference. UniversalVTG pursues an orthogonal, highly efficient route: a model over two orders of magnitude smaller. By unifying cross-dataset supervision via standardized query representations and large-scale pretraining, our single compact architecture achieves strong generalization across diverse benchmarks without the massive compute overhead of MLLM pipelines.

\section{Preliminaries}
We briefly formalize video temporal grounding (VTG) and summarize the standard modeling pipeline, to clarify the components we build upon.

\subsection{Problem Setup}
Given an untrimmed video $\mathcal{V}$ and a natural-language \emph{text input} $\mathcal{Q}$—which may appear as a keyword phrase, a declarative description, or a question—VTG localizes the temporal segment $(t_s,t_e)$ that semantically matches the query. Let $\mathcal{V}=\{f_i\}_{i=1}^{N_f}$ denote sampled frames (or clips) with timestamps $\mathcal{T}=\{t_i\}_{i=1}^{N_f}$. The task is
\begin{equation}
(t_s,t_e) = \Phi(\mathcal{V},\mathcal{Q}), \qquad t_s,t_e \in \mathcal{T}.
\end{equation}
We follow the standard benchmark setting where each $\mathcal{Q}$ corresponds to a single contiguous moment.

\subsection{Standard VTG Pipeline and Implications}
Most approaches decompose VTG into (i) \emph{feature extraction} and (ii) \emph{temporal grounding}.
A visual encoder (e.g., EgoVLP~\cite{lin2022egocentric}, InternVideo~\cite{wang2022internvideo}, CLIP~\cite{radford2021learning}) maps the video to a temporal feature sequence $V=\{v_i\}_{i=1}^{L_V}\in\mathbb{R}^{L_V\times D_v}$, while a text encoder maps the query to token features $Q=\{q_j\}_{j=1}^{L_Q}\in\mathbb{R}^{L_Q\times D_q}$.
A VTG model then fuses $V$ and $Q$ and predicts $(t_s,t_e)$.

Two structural observations follow from this formulation.
First, cross-dataset training implicitly assumes that videos and text from different benchmarks are embedded into a compatible representation space.
Second, for long untrimmed videos, feature extraction typically dominates computational cost~\cite{ramakrishnan2023spotem}, making the efficiency of the visual encoder a primary bottleneck. These observations highlight that any universal VTG system must (i) operate within a shared video-text representation space across datasets, and (ii) remain computationally scalable for long-video processing.

\begin{figure}[t!]
    \centering
    \includegraphics[width=\linewidth]{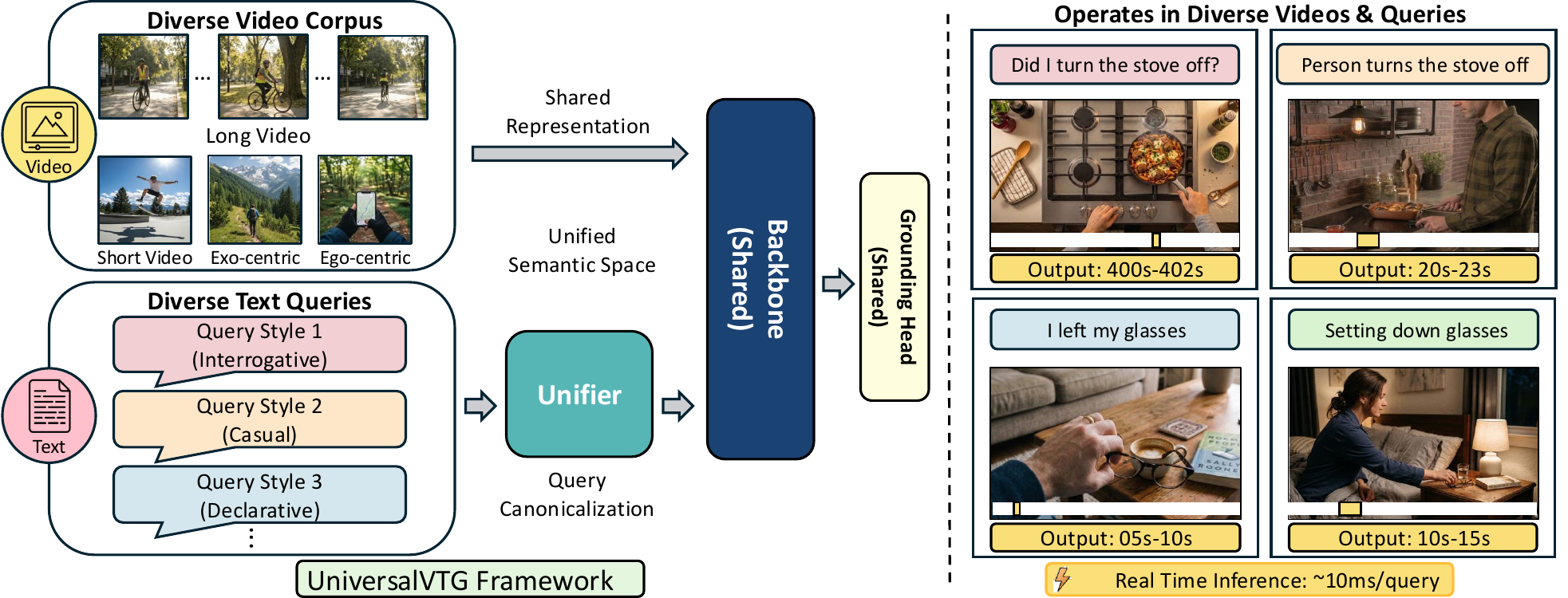}
    \caption{\textbf{UniversalVTG Framework.} A single, lightweight model generalizes across heterogeneous video domains and query styles. (Left) Diverse videos are mapped to a \emph{Shared Representation} via an efficient backbone, while a \emph{Query Unifier} canonicalizes multi-style text inputs into a \emph{Unified Semantic Space}. (Right) With both modalities standardized, one grounding head localizes events across varied viewpoints (ego/exo), durations (short/long), and linguistic forms (e.g., questions, declarations). UniversalVTG achieves real-time inference ($\sim$10\,ms/query) suitable for long-form deployment.}
    \label{fig:main}
    \vspace{-5mm}
\end{figure}

\section{Method}
Our goal is to develop a single VTG model that generalizes across heterogeneous datasets and query styles while remaining computationally practical for real-world deployment. UniversalVTG follows the standard VTG decomposition—feature extraction followed by a temporal grounding head—but aligns cross-dataset supervision by canonicalizing queries into a shared format. UniversalVTG comprises: (i) a single efficient visual encoder shared across datasets (Section~\ref{sec:backbone}); (ii) an offline \emph{unifier} that maps heterogeneous query formulations into a consistent instruction template, mitigating negative transfer (Section~\ref{sec:datasets}); and (iii) a lightweight temporal grounding head (Section~\ref{sec:vtg_head}) strengthened through large-scale cross-dataset pretraining (Section~\ref{sec:pretraining}). Together, these design choices yield a universal foundation model for grounding that can be trained once and deployed seamlessly across diverse benchmarks.

\vspace{-2mm}
\subsection{Visual Encoder}
\label{sec:backbone}
\vspace{-1mm}
The choice of visual encoder largely determines both grounding accuracy and computational cost. Since the grounding head can only operate on the visual encoder's extracted features—and feature extraction dominates runtime for long videos—we treat backbone selection as a first-order design decision.

Most VTG systems rely on \emph{video}-based encoders (e.g., SlowFast~\cite{feichtenhofer2019slowfast}, TimeSFormer~\cite{bertasius2021space}, C3D~\cite{tran2015learning}) or large-scale video foundation models (e.g., InternVideo~\cite{wang2022internvideo}, EgoVLP~\cite{lin2022egocentric}). These clip-based models explicitly encode motion and often achieve strong performance, but introduce two obstacles for a universal model: (i) backbone choices are typically dataset-specific (e.g., egocentric vs.\ third-person pretraining), fragmenting the representation space; and (ii) clip-based processing substantially increases token count and compute, with feature extraction often dominating VTG cost~\cite{ramakrishnan2023spotem}.

An alternative is to use \emph{image}-based encoders (e.g., CLIP~\cite{radford2021learning}), extracting sparse frames at lower cost. While commonly viewed as less temporally expressive, we hypothesize that strong per-frame semantics—if trained to transfer across domains—are sufficient for VTG. This motivates seeking a lightweight, broadly transferable image-based backbone rather than defaulting to expensive video-native models.

We evaluate candidate backbones to quantify the accuracy–efficiency trade-off (Table~\ref{tab:backbone}). InternVideo achieves the highest accuracy (29.0 Avg.\ R@1\&5) but at 161.0 TFLOPs/min. SlowFast and CLIP-ViT-L/14 are cheaper (7.40 and 21.0 TFLOPs/min) but substantially weaker. \textbf{PerceptionEncoder} offers the best balance: 26.6 Avg.\ R@1\&5 at 21.1 TFLOPs/min, yielding a \(\sim7.6\times\) reduction in compute relative to InternVideo while approaching its accuracy. This identifies an efficient design point for VTG feature extraction. We therefore adopt \textbf{PerceptionEncoder} as UniversalVTG’s shared visual encoder, standardizing representations across datasets while keeping long-video processing tractable.

\vspace{-5mm}
\begin{table}[H]
\centering
\caption{\textbf{Backbone comparison on Ego4D-NLQ.}
We report feature extraction rate (FPS), mean of R@1\&5 (evaluated at IoU 0.3/0.5), and feature-extraction compute (TFLOPs/min).
We follow prior work~\cite{an2025hieramamba} and use the same extraction settings per backbone for a fair comparison:
\emph{video-based} encoders process 32-frame clips with stride 16 on 30\,FPS videos, whereas \emph{image-based} encoders operate at 2 frames per second.}
\vspace{-2mm}
\label{tab:backbone}
\resizebox{0.95\linewidth}{!}{%
\setlength{\tabcolsep}{7pt}
\begin{tabular}{@{}l ccc@{}}
\toprule
\textbf{Backbone} 
& \makecell{\textbf{Feature}  \textbf{FPS}} 
& \makecell{\textbf{Avg.}  \textbf{R@1\&5} \textbf{($\uparrow$)}} 
& \makecell{\textbf{TFLOPs}  \textbf{/ min. video} \textbf{($\downarrow$)}} \\
\midrule
\multicolumn{4}{c}{\textit{Video-based backbones}} \\
\midrule
SlowFast~\cite{feichtenhofer2019slowfast}      & 1.88 & 16.3 & 7.40 \\
EgoVLP~\cite{lin2022egocentric}        & 1.88 & 25.7 & 83.1 \\
InternVideo~\cite{wang2022internvideo}   & 1.88 & 29.0 & 161.0 \\
\midrule
\multicolumn{4}{c}{\textit{Image-based backbones}} \\
\midrule
CLIP-ViT-L14~\cite{radford2021learning}  & 2.00 & 20.0 & 21.0 \\
\rowcolor{gray!12} PerceptionEncoder-L~\cite{bolya2025perception} & 2.00 & 26.6 & 21.1 \\
\bottomrule
\end{tabular}%
}
\end{table}

\begin{table}[t]
\centering

\begin{minipage}[c]{0.45\textwidth}
\centering
\caption{Cross-dataset evaluation (AVG R@1). Rows indicate the \textbf{training} dataset and columns indicate the \textbf{evaluation} dataset. Dataset-specific models transfer poorly to unseen domains.}
\label{tab:cross_dataset}
\resizebox{\linewidth}{!}{%
\begin{tabular}{@{}l cccc@{}}
\toprule
\makecell[l]{\textbf{Test $\rightarrow$} \\ \textbf{Train $\downarrow$}} 
& \makecell{\textbf{NLQ}} 
& \makecell{\textbf{TACoS}} 
& \makecell{\textbf{Charades} \\ \textbf{-STA}} 
& \makecell{\textbf{ANet} \\ \textbf{-Cap.}} \\
\midrule
NLQ          & \textbf{19.55} & 20.18 & 23.50 & 13.16 \\
TACoS        & 3.39 & \textbf{56.70} & 13.23 & 10.78 \\
Cha.-STA     & 4.35 & 12.92 & \textbf{62.35} & 9.35 \\
ANet-Cap.    & 3.93 & 13.97 & 22.43 & \textbf{39.52} \\
\bottomrule
\end{tabular}%
}
\end{minipage}\hfill
\begin{minipage}[c]{0.52\textwidth}
\centering
\includegraphics[width=\linewidth]{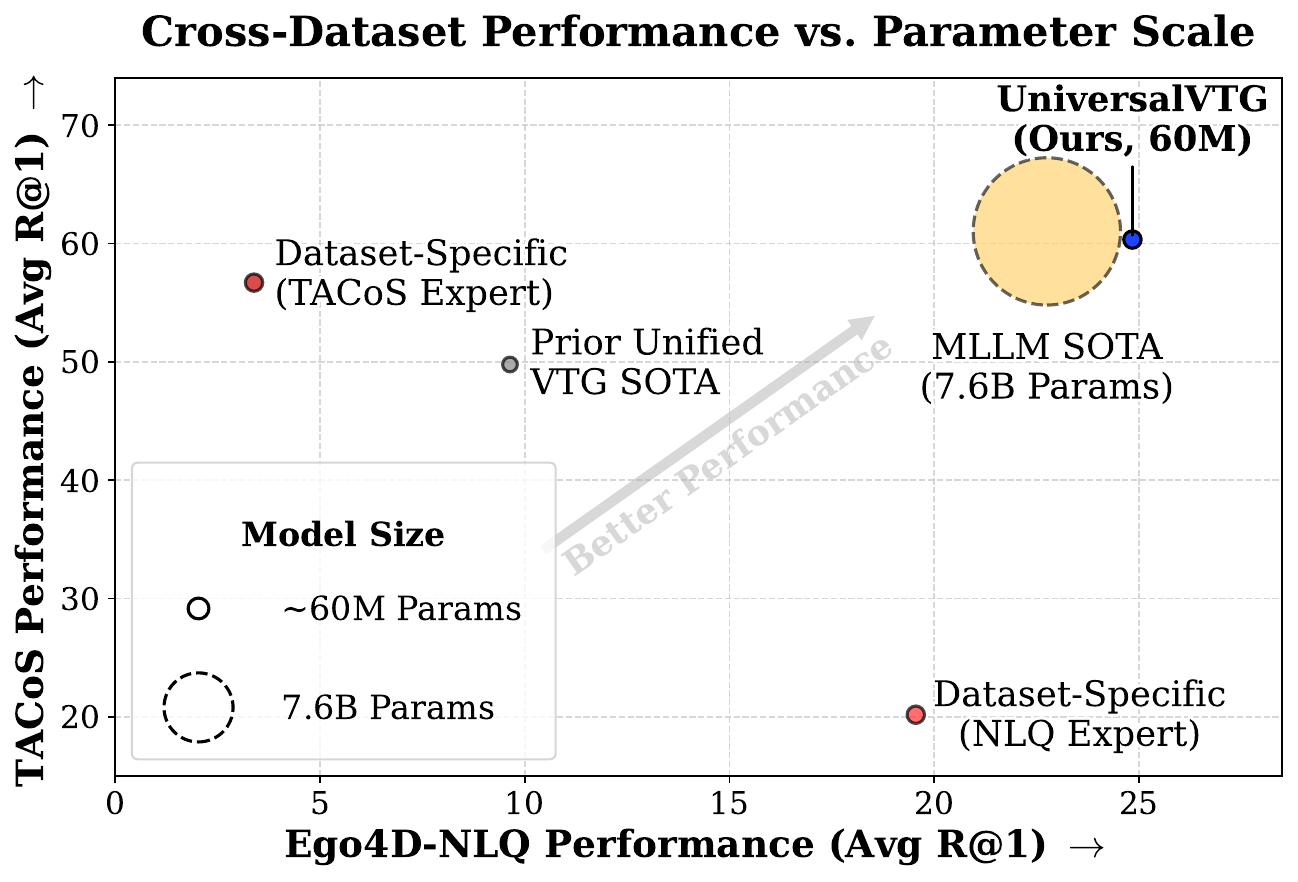}
\captionof{figure}{Cross-Dataset Performance vs. Parameter Scale. \textbf{UniversalVTG} achieves performance parity with $100\times$ larger MLLMs while outperforming specialized expert models.}
\label{fig:efficiency_bubble}
\end{minipage}
\vspace{-7mm}
\end{table}

\subsection{Cross-Dataset Training: Diagnosing and Mitigating Negative Transfer}
\label{sec:datasets}
\vspace{-2mm}
Adopting a shared visual encoder projects videos from heterogeneous benchmarks into a common feature space. 
To study whether a single VTG model can generalize across diverse grounding regimes, we evaluate on five widely adopted benchmarks: GoalStep~\cite{song2023ego4d}, Ego4D-NLQ~\cite{grauman2022ego4d}, TACoS~\cite{regneri2013grounding}, Charades-STA~\cite{sigurdsson2016hollywood}, and ActivityNet-Captions~\cite{krishna2017dense}. We chose this set to provide 
comprehensive coverage of the dominant sources of cross-dataset mismatch in VTG: egocentric~\cite{song2023ego4d, grauman2022ego4d} vs.\ third-person~\cite{regneri2013grounding, krishna2017dense, sigurdsson2016hollywood} viewpoints, short-form~\cite{krishna2017dense,sigurdsson2016hollywood} vs.\ long-form~\cite{song2023ego4d,grauman2022ego4d,regneri2013grounding} videos (average duration ranging from tens of seconds to nearly half an hour), and diverse query/annotation conventions spanning different styles and densities. Together, they form a representative cross-section of contemporary VTG settings for evaluating unified training under substantial heterogeneity (Table~\ref{tab:dataset_stats}).

Crucially, this extreme heterogeneity severely penalizes standard dataset-specific training. As shown in Table~\ref{tab:cross_dataset}, a model optimized for one domain transfers exceptionally poorly to unseen settings, suffering catastrophic performance drops. This fragmentation motivates training a single model across all domains. While visual alignment removes an architectural barrier to such unification, it does not guarantee effective joint training. In practice, simply aggregating supervision across these heterogeneous datasets introduces negative transfer (Table~\ref{tab:unified_vs_ft}). We analyze this behavior and propose strategies to unify language supervision.

\vspace{-2mm}
\subsubsection{Naïve Joint Training.}

We fix the visual encoder (Section~\ref{sec:backbone}) and 
adopt a standard lightweight temporal grounding head (HieraMamba~\cite{an2025hieramamba}, detailed in Section~\ref{sec:vtg_head}).
We then compare two training regimes: (i) \textbf{dataset-specific training}, where separate models are optimized independently for each dataset, and (ii) \textbf{naïve joint training}, where a single model is trained on the union of five VTG benchmarks. As shown in Table~\ref{tab:unified_vs_ft}, despite identical architectures, naïve joint training underperforms dataset-specific training on most benchmarks, with drops of up to 3.56\%. Ego4D-NLQ is a mild exception, showing a small gain; however, overall the results indicate negative transfer arising from cross-dataset heterogeneity.

\begin{table}[H]
\centering
\caption{\textbf{Effect of joint training and language unification.}
All settings use the same visual encoder and grounding head (HieraMamba); only the training regime differs.
Performance is reported as Avg. R@1 (higher is better).
$\Delta$ denotes relative change compared to dataset-specific training.}
\label{tab:unified_vs_ft}
\resizebox{\linewidth}{!}{%
\begin{tabular}{l cc cc cc cc cc}
\toprule
\multirow{2}{*}{\textbf{Training Regime}} 
& \multicolumn{2}{c}{\textbf{GoalStep}} 
& \multicolumn{2}{c}{\textbf{NLQ}} 
& \multicolumn{2}{c}{\textbf{TACoS}} 
& \multicolumn{2}{c}{\textbf{Charades-STA}} 
& \multicolumn{2}{c}{\textbf{ANet-Cap.}} \\
\cmidrule(lr){2-3} \cmidrule(lr){4-5} \cmidrule(lr){6-7} \cmidrule(lr){8-9} \cmidrule(lr){10-11}
& Perf. & $\Delta$ 
& Perf. & $\Delta$ 
& Perf. & $\Delta$ 
& Perf. & $\Delta$ 
& Perf. & $\Delta$ \\
\midrule
Dataset-Specific & 22.92 & -- & 19.55 & -- & 56.70 & -- & 62.35 & -- & 39.52 & -- \\
Joint (Naïve) & 22.22 & -3.05\% & 20.16 & +3.09\% & 55.16 & -2.72\% & 61.68 & -1.07\% & 38.11 & -3.56\% \\
\rowcolor{gray!12}
\textbf{Joint (Unified)} 
& \textbf{23.10} & \textbf{+3.94\%} 
& \textbf{21.14} & \textbf{+4.86\%} 
& \textbf{58.76} & \textbf{+6.53\%} 
& \textbf{63.25} & \textbf{+2.55\%} 
& \textbf{40.14} & \textbf{+5.31\%} \\
\bottomrule
\end{tabular}}
\end{table}


\subsubsection{Diagnosis: Query Formulation Mismatch.}

We attribute much of this gap to heterogeneity in language supervision. In addition to visual domain shifts (e.g., egocentric versus third-person video), VTG datasets differ substantially in how grounding intent is expressed. Queries may take the form of interrogative questions (e.g., ``What did the person do before opening the fridge?''), declarative captions (``The person opens the fridge''), short imperative fragments (``open fridge''), or longer narrative descriptions. This stylistic variation induces a fragmented language distribution. With compact text encoders (e.g., CLIP-style pooled representations) and lightweight cross-modal fusion, the model has limited capacity to normalize heterogeneous query styles. As a result, it must learn temporal localization while also becoming invariant to query formulation. When datasets are merged, this added burden increases optimization difficulty and amplifies negative transfer.

\subsubsection{Remedy: Semantic Canonicalization via Query Unification.}
\label{sec:query_unifier}
To mitigate cross-benchmark linguistic friction, we standardize language supervision at the source. As shown in Fig.~\ref{fig:main}, we introduce a \textbf{Query Unifier} that maps dataset-specific query distributions into a shared \emph{declarative} canonical space \emph{offline}. Concretely, we use an instruction-tuned LLM~\cite{yang2025qwen3, singh2025openai, grattafiori2024llama} to harmonize tense and grammatical structure across the training corpus, converting heterogeneous inputs into a consistent past-tense format (Table~\ref{tab:query_examples}). For example, the interrogative \textit{``What did I put in the bucket?''} (Ego4D-NLQ) is mapped to \textit{``I placed an object into the bucket.''} We analyze unifier choices (models/prompts) and their efficiency trade-offs in the Supplementary.

This canonicalization removes the need for the grounding model to implicitly normalize stylistic disparities during joint training. The decoder can therefore devote capacity to the core objective—aligning the canonicalized semantic intent to its corresponding video segment—rather than learning dataset-specific query conventions. Empirically, unified joint training consistently outperforms naïve joint training and dataset-specific training across all five benchmarks (Table~\ref{tab:unified_vs_ft}). Crucially, by resolving this linguistic friction at the data level, our lightweight architecture achieves cross-dataset generalization comparable to multi-billion parameter MLLMs (Fig.~\ref{fig:efficiency_bubble}).

\vspace{-3mm}
\begin{table}[H]
\centering
\caption{Qualitative examples of semantic canonicalization. The Query Unifier maps dataset-specific query styles (questions, fragments, tense/aspect mismatches) into a standardized declarative format used for joint training.}
\label{tab:query_examples}
\resizebox{\columnwidth}{!}{%
\setlength{\tabcolsep}{5.5pt}
\begin{tabular}{@{}llll@{}}
\toprule
\textbf{Benchmark} & \textbf{Original Query (Input)} & \textbf{Unified Query (Canonical Space)} & \textbf{Resolved Friction} \\
\midrule
\textsc{GoalStep} & \textit{``Add onion to the pan''} & \textit{``A person added onion to the pan''} & Imperative, No Subject \\
\textsc{Ego4D-NLQ} & \textit{``What did I put in the bucket?''} & \textit{``I placed an object into the bucket.''} & Interrogative\\
\textsc{TACoS} & \textit{``Takes a cup out of the cabinet.''} & \textit{``A person took a cup out of the cabinet.''} & Missing Subject \\
\textsc{Charades-STA} & \textit{``person takes a cup out the fridge.''} & \textit{``A person took a cup out of the fridge.''} & Missing Preposition  \\
\textsc{ActivityNet} & \textit{``A woman is walking along a track.''} & \textit{``A woman walked along a track.''} & Present Continuous \\
\bottomrule
\end{tabular}%
}
\vspace{-3mm}
\end{table}

\subsection{Temporal Grounding Architecture}
\label{sec:vtg_head}

With unified visual and language representations in place, we require a temporal decoder that scales to long sequences. Modern lightweight VTG architectures typically follow a standard paradigm: independent unimodal encoding followed by cross-modal fusion and boundary prediction. We adopt HieraMamba~\cite{an2025hieramamba} as our instantiation of this paradigm. Because it utilizes state-space models rather than standard quadratic-time transformers, it efficiently scales to long video contexts while maintaining the standard generic frame/clip feature interface.

Importantly, because HieraMamba is highly representative of this broader class of fusion-based lightweight decoders, our findings are broadly applicable. Our core contribution, query canonicalization, operates strictly at the data and representation interface. It can therefore be plugged into other standard grounding decoders without architectural modification. Frame-level visual features are simply fused with the canonicalized and enriched text features within this architecture.

Training follows the standard multi-task objective used in HieraMamba, combining classification and boundary regression losses:
\begin{equation}
\mathcal{L}_{total} = \lambda_{cls}\mathcal{L}_{cls} + \lambda_{reg}\mathcal{L}_{reg} + \mathcal{L}_{contrast},
\end{equation}
where $\mathcal{L}_{cls}$ is Focal Loss~\cite{lin2017focal} and $\mathcal{L}_{reg}$ is Distance-IoU (DIoU) loss~\cite{zheng2020distance}. We retain HieraMamba’s contrastive objectives ($\mathcal{L}_{ACC}$ and $\mathcal{L}_{SPC}$) and default loss weights~\cite{an2025hieramamba}, ensuring that performance gains stem from unified supervision rather than architectural modifications.

\begin{table}[t]
\centering
\caption{
Statistics of the video temporal grounding datasets used in this work. We partition the datasets into our large-scale Stage-I pretraining corpus (top) and the five Stage-II target benchmarks used for joint fine-tuning (bottom). Notably, our Stage-I corpus aggregates over 1M query--segment pairs, providing an order of magnitude more temporal supervision than the combined Stage-II training datasets.
}
\label{tab:dataset_stats}
\resizebox{\linewidth}{!}{%
\setlength{\tabcolsep}{5.5pt}
\begin{tabular}{@{}l rrrr l@{}}
\toprule
\multicolumn{1}{@{}c}{\textbf{Dataset}} 
& \multicolumn{1}{c}{\makecell{\textbf{\#} \textbf{Videos}}} 
& \multicolumn{1}{c}{\makecell{\textbf{\#} \textbf{Queries}}} 
& \multicolumn{1}{c}{\makecell{\textbf{Avg. Video} \\ \textbf{Length (s)}}} 
& \multicolumn{1}{c}{\makecell{\textbf{Avg. Seg.} \\ \textbf{Length (s)}}} 
& \multicolumn{1}{c@{}}{\textbf{Domain}} \\
\midrule

\multicolumn{6}{c}{\textit{Stage-I: Pretraining Datasets}} \\
\midrule
NaQ~\cite{ramakrishnan2023naq}             & 5,018 & 858,350 & 416 & 9.7  & Open (Ego) \\
Momentor~\cite{qian2024momentor}        & 9,697  & 250,738 & 408 & 3.64 & Open \\
COIN~\cite{tang2019coin}            & 7,938  & 31,251 & 146 & 14.9 & Open \\
YouCook2~\cite{zhou2018towards}        & 1,560  & 12,066 & 316 & 19.6 & Cooking \\
HiREST~\cite{zala2023hierarchical}          & 583    & 4,414  & 268 & 19.2 & Open \\
\midrule

\multicolumn{6}{c}{\textit{Stage-II: Target Temporal Grounding Benchmarks (train/val)}} \\
\midrule
GoalStep~\cite{song2023ego4d}        & 583/134 & 31,566/7,696 & 1,500/1674 & 32.2/35.4 & Instructional \\
Ego4D-NLQ~\cite{grauman2022ego4d}       & 1,270/415 & 13,847/4,552 & 494/500 & 11.3/10.8 & Episodic \\
TACoS~\cite{regneri2013grounding}           & 75/25 & 9,790/4,001 & 224/368 & 23.3/31.9 & Cooking \\
Charades-STA~\cite{sigurdsson2016hollywood}    & 5,338/1,334 & 12,408/3,720 & 31/30 & 8.3/8.0 & Indoor Daily \\
ActivityNet-Cap.~\cite{krishna2017dense} & 10,002/4,872 & 37,399/16,990 & 117/118 & 35.5/40.3 & Open (YouTube) \\
\bottomrule
\end{tabular}%
}
\end{table}

\subsection{Large-Scale Cross-Dataset Pretraining}
\label{sec:pretraining}

In standard VTG pipelines, visual features are extracted using pretrained vision--language encoders (e.g., CLIP~\cite{radford2021learning} or EgoVLP~\cite{lin2022egocentric}), while the temporal grounding head is trained from scratch on a single target dataset. As a result, the grounding head learns cross-modal temporal alignment from a relatively narrow distribution, which can limit robustness to new domains and query formulations.

Because UniversalVTG uses a shared visual representation (Section~\ref{sec:backbone}) and canonicalizes language supervision across datasets (Section~\ref{sec:datasets}), we can aggregate heterogeneous grounding datasets and pretrain the temporal grounding head at scale. Concretely, Stage-I pretraining uses over one million query--segment pairs (Table~\ref{tab:dataset_stats}), providing a strong VTG-specific initialization for subsequent joint fine-tuning and improving generalization across benchmarks.

\subsubsection{Pretraining Data Curation and Scale.}
To build a comprehensive foundation for universal video temporal grounding, we establish strict inclusion criteria for our pretraining corpus. Rather than arbitrarily combining datasets, we specifically aggregate publicly available sources that satisfy three core requirements: (1) they provide precise, temporally anchored language annotations (explicit start and end boundaries); (2) they span the full spectrum of visual perspectives necessary for a universal model (both first-person egocentric and third-person exocentric); and (3) they feature complex, untrimmed videos that demand fine-grained procedural or temporal reasoning. 
Applying these criteria yields a definitive collection of five large-scale datasets for our Stage-I pretraining: NaQ~\cite{ramakrishnan2023naq}, Momentor~\cite{qian2024momentor}, COIN~\cite{tang2019coin}, YouCook2~\cite{zhou2018towards}, and HiREST~\cite{zala2023hierarchical} (Table~\ref{tab:dataset_stats}). Together, this curated selection systematically covers our required domain variations, including egocentric daily routines (NaQ), third-person instructional content (COIN, YouCook2), hierarchical procedures (HiREST), and fine-grained temporal reasoning scenarios (Momentor). 
For datasets distributed via video URLs, we retain only videos that were successfully accessible at the time of collection. As detailed in Table~\ref{tab:dataset_stats}, the resulting Stage-I pretraining set aggregates over 1.16 million query--segment pairs. This provides an order of magnitude more supervision than the combined Stage-II target benchmarks, yielding the massive, task-specific scale necessary to drive cross-domain generalization.

\subsubsection{Unified Training Protocol}

UniversalVTG is trained in two stages. In Stage-I, we pretrain the grounding head on the aggregated canonicalized corpus while keeping the visual encoder frozen, enabling the model to learn a shared cross-modal alignment space without perturbing the visual representation. In Stage-II, we initialize from the Stage-I checkpoint and jointly fine-tune on all target VTG benchmarks using a single shared model, with no dataset-specific heads or hyperparameter tuning. This two-stage strategy stabilizes optimization and allows heterogeneous supervision to be absorbed into a unified grounding model.

\section{Experiments}
\label{sec:experiments}

To thoroughly evaluate UniversalVTG, we conduct comprehensive experiments across diverse video domains and sequence lengths, comparing our single unified model against both dataset-specific VTG architectures and recent parameter-heavy MLLMs.

\subsection{Experimental Setup}

\subsubsection{Datasets.}

We evaluate on the same five VTG benchmarks introduced in Section~\ref{sec:datasets} (see Table~\ref{tab:dataset_stats} for dataset statistics and Section~\ref{sec:datasets} for the selection rationale). For clarity, we group them by video duration into long-form and shorter-form benchmarks:

\noindent\textbf{Long-form benchmarks.}
Ego4D-NLQ~\cite{grauman2022ego4d} (egocentric episodic videos), TACoS~\cite{regneri2013grounding} (third-person cooking), and GoalStep-StepGrounding~\cite{song2023ego4d} (long instructional videos; avg.\ $\sim$26 minutes).

\noindent\textbf{Shorter-form benchmarks.}
Charades-STA~\cite{sigurdsson2016hollywood} (indoor daily activities) and ActivityNet-Captions~\cite{krishna2017dense} (open-domain YouTube videos).

\vspace{-2mm}
\subsubsection{Evaluation Metrics.}
Following standard VTG protocols, we report Recall@$K$ at temporal Intersection-over-Union (IoU) thresholds, denoted as R@$K$@IoU=$\theta$. We adopt the benchmark-specific evaluation conventions used in prior work: IoU $\in \{0.3,0.5\}$ for the long-form datasets (GoalStep, Ego4D-NLQ, TACoS) and IoU $\in \{0.5,0.7\}$ for the short-form datasets (Charades-STA, ActivityNet-Captions). In the main text, we emphasize Recall@1 to enable direct comparison with recent MLLM-based methods, which often report only top-1 localization. We provide the full Recall@5 results in the Supplementary Material to maintain continuity with the broader VTG literature.

\vspace{-2mm}
\subsubsection{Implementation Details.}
We use Perception Encoder-L (\texttt{\small PE-Core-L14\allowbreak-\allowbreak336})~\cite{bolya2025perception} as a shared visual backbone and extract frame-level features offline at 2~FPS. Unless otherwise noted, the visual encoder is frozen, and the resulting temporal feature sequences are processed by the HieraMamba~\cite{an2025hieramamba} grounding head.

\textsc{UniversalVTG} is trained in two stages. In Stage-I (pretraining), we use 8 GPUs with a learning rate of $1\times10^{-3}$. Each iteration constructs a balanced batch by sampling, \emph{from each pretraining dataset}, 8 videos with 2 queries per video per GPU. In Stage-II (joint fine-tuning), we fine-tune a single shared model on 1 GPU with a learning rate of $1\times10^{-4}$. Each iteration similarly samples, \emph{from each target dataset}, 4 videos and 2 queries per video, jointly optimizing across all five benchmarks.

\begin{table*}[t]
\centering
\caption{Comparison across five VTG benchmarks. R@1 is reported at multiple IoU thresholds with the average shown in each block. Models above UniversalVTG are dedicated VTG architectures; models below are MLLM-based approaches. ``--'' indicates results not reported by the original authors (thus unavailable for direct comparison). Models marked with $\blacksquare$ utilize shared weights across all benchmarks, whereas $\square$ denotes models requiring separate weights per dataset.}
\label{tab:main_results}

\resizebox{\textwidth}{!}{%
\begin{tabular}{@{}lc ccc ccc ccc ccc ccc@{}}
\toprule
\multirow{3}{*}{\textbf{Method}} & \multirow{3}{*}{\makecell{\textbf{Shared} \\ \textbf{Weights}}}
& \multicolumn{9}{c}{\textbf{Long-Form Videos}}
& \multicolumn{6}{c}{\textbf{Short-Form Videos}} \\
\cmidrule(lr){3-11} \cmidrule(lr){12-17}
& & \multicolumn{3}{c}{\textbf{GoalStep}}
& \multicolumn{3}{c}{\textbf{Ego4D-NLQ}}
& \multicolumn{3}{c}{\textbf{TACoS}}
& \multicolumn{3}{c}{\textbf{Cha.-STA}}
& \multicolumn{3}{c}{\textbf{ANet-Cap.}} \\
\cmidrule(lr){3-5} \cmidrule(lr){6-8} \cmidrule(lr){9-11} \cmidrule(lr){12-14} \cmidrule(lr){15-17}
& & 0.3 & 0.5 & Avg.
  & 0.3 & 0.5 & Avg.
  & 0.3 & 0.5 & Avg.
  & 0.5 & 0.7 & Avg.
  & 0.5 & 0.7 & Avg. \\
\midrule

\multicolumn{17}{c}{\textit{Dedicated VTG Models}} \\
\midrule
2D-TAN~\cite{zhang2020learning}         & $\square$ & -- & -- & -- & 5.04 & 2.05 & 3.55 & 45.61 & 35.77 & 40.69 & 56.64 & 36.21 & 46.43 & 46.16 & 29.21 & 37.69 \\
M-DETR~\cite{lei2021detecting}         & $\square$ & -- & -- & -- & 8.23 & 5.01 & 6.62 & 37.97 & 24.67 & 31.32 & 52.07 & 30.59 & 41.33 & -- & -- & -- \\
BAM-DETR~\cite{lee2024bam}       & $\square$ & -- & -- & -- & -- & -- & -- & 56.69 & 41.54 & 49.12 & 59.95 & 39.38 & 49.67 & -- & -- & -- \\
CONE~\cite{hou2022cone}           & $\square$ & -- & -- & -- & 14.15 & 8.18 & 11.17 & -- & -- & -- & -- & -- & -- & -- & -- & -- \\
UnLoc-L~\cite{yan2023unloc}        & $\square$ & -- & -- & -- & -- & -- & -- & -- & -- & -- & 60.80 & 38.40 & 49.60 & 48.30 & 30.20 & 39.25 \\
SnAG~\cite{snag}           & $\square$ & -- & -- & -- & 15.72 & 10.78 & 13.25 & 56.44 & 44.86 & 50.65 & 51.72 & 33.52 & 42.62 & 48.55 & \textbf{30.56} & \textbf{39.56} \\
DeCafNet~\cite{lu2025decafnet}       & $\square$ & 21.29 & 17.46 & 19.38 & 18.10 & 12.55 & 15.33 & 57.36 & 46.79 & 52.08 & 68.79 & 47.55 & 58.17 & -- & -- & -- \\
HieraMamba~\cite{an2025hieramamba} & $\square$ & -- & -- & -- & 18.81 & 13.04 & 15.93 & 59.59 & 48.99 & 54.29 & -- & -- & -- & -- & -- & -- \\
UniVTG~\cite{lin2023univtg}         & $\blacksquare$  & -- & -- & -- & 11.74 & 7.54 & 9.64 & 56.11 & 43.44 & 49.78 & 60.19 & 38.55 & 49.37 & 42.41 & 21.55 & 31.98 \\
\midrule
\rowcolor{gray!15} \textbf{UniversalVTG} & $\blacksquare$
& \textbf{26.75} & \textbf{22.12} & \textbf{24.44}
& \textbf{29.06} & \textbf{20.56} & \textbf{24.81}
& \textbf{64.71} & \textbf{53.11} & \textbf{58.91}
& \textbf{73.87} & \textbf{56.42} & \textbf{65.15}
& \textbf{48.98} & 29.79 & 39.39 \\
\midrule

\multicolumn{17}{c}{\textit{MLLM-based Methods}} \\
\midrule
Qwen2.5-VL-7B~\cite{bai2025qwen3}     & $\blacksquare$ & -- & -- & -- & 1.11 & 0.48 & 0.80 & 7.66 & 3.35 & 5.51 & 60.32 & 34.27 & 47.30 & 16.96 & 8.75 & 12.86 \\
TimeChat-7B~\cite{ren2024timechat}       & $\blacksquare$ & -- & -- & -- & -- & -- & -- & 27.70 & 15.10 & 21.40 & 46.70 & 23.70 & 35.20 & -- & -- & -- \\
HawkEye-7B~\cite{wang2024hawkeye}        & $\blacksquare$ & -- & -- & -- & -- & -- & -- & -- & -- & -- & 58.30 & 28.80 & 43.55 & 34.70 & 17.90 & 26.30 \\
ED-VTG-7B~\cite{pramanick2025enrich}         & $\blacksquare$ & -- & -- & -- & -- & -- & -- & 46.00 & 31.50 & 38.75 & 62.10 & 35.00 & 48.55 & 45.10 & 22.70 & 33.90 \\
UniTime-2B~\cite{li2025universal}     & $\blacksquare$ & -- & -- & -- & 10.50 & 5.80 & 8.15 & -- & -- & -- & 65.38 & 42.18 & 53.78 & -- & -- & -- \\
UniTime-7B~\cite{li2025universal}   & $\blacksquare$ & -- & -- & -- & 27.09 & 18.41 & 22.75 & 66.91 & 55.14 & 61.03 & 75.27 & 56.85 & 66.06 & 53.67 & 35.90 & 44.79 \\
\bottomrule
\end{tabular}%
}
\end{table*}

\subsection{Comparison with State-of-the-Art VTG Models}
\label{sec:sota_vtg}
We compare \textsc{UniversalVTG} to dedicated \textsc{vtg} architectures and \textsc{mllm}-based approaches across five benchmarks (Table~\ref{tab:main_results}). Fig.~\ref{fig:teaser} highlights the main result: with one shared checkpoint, \textsc{UniversalVTG} matches or exceeds dataset-specific experts trained separately per benchmark and improves over prior unified models.

\noindent\textit{Results across regimes.}
As shown in Table~\ref{tab:main_results}, \textsc{UniversalVTG} is consistently strong on both long-form benchmarks (GoalStep, Ego4D-NLQ, TACoS) and shorter-form benchmarks (Charades-STA, ActivityNet-Captions) using a single shared checkpoint. Overall, it matches or surpasses dataset-specific experts that require separate per-dataset training, and improves over prior unified model~\cite{lin2023univtg} (Fig.~\ref{fig:teaser}). In practice, this eliminates the need to train, tune, and maintain five separate dataset-specific checkpoints—reducing training and deployment overhead by a factor of $5\times$ and enabling a single plug-and-play model across benchmarks.

\subsection{Comparison with MLLM-based Approaches}
\label{sec:mllm_comparison}

Table~\ref{tab:main_results} shows that \textsc{UniversalVTG} consistently outperforms MLLM-based VTG systems such as \textit{Qwen2.5-VL-7B}~\cite{bai2025qwen3}, \textit{TimeChat}, \textit{HawkEye}, and \textit{ED-VTG} across the benchmarks where they report results, despite using only 60M parameters. Against the strongest MLLM baseline, \textit{UniTime}, \textsc{UniversalVTG} exceeds the 2B variant by a large margin and remains competitive with the 7B variant while being over $100\times$ smaller.

The advantage is most pronounced in long-video regimes. Many MLLM-based methods omit \textsc{GoalStep} due to context-window and memory constraints on 26-minute videos, whereas \textsc{UniversalVTG} directly processes these untrimmed sequences and achieves 24.44 Avg.\ R@1 on \textsc{GoalStep}.

Efficiency-wise, \textsc{UniversalVTG} provides comparable accuracy with orders-of-magnitude lower grounding cost: as shown in Table~\ref{tab:efficiency_grounder}, it reduces grounding parameters, TFLOPs, and runtime dramatically relative to \textit{UniTime}-7B. Fig.~\ref{fig:efficiency_bubble} provides a visual summary of this trade-off on a representative cross-dataset pairing (additional pairings in supplementary).

\begin{table}[t]
\centering
\caption{\textbf{Grounding-head efficiency comparison.}
We report parameter count, computational cost (TFLOPs), and inference runtime for the grounding module, measured on a 500\,s video (average Ego4D-NLQ length) with a single NVIDIA A40 GPU.\protect\footnotemark}
\label{tab:efficiency_grounder}

\resizebox{0.85\linewidth}{!}{%
\setlength{\tabcolsep}{20pt}
\begin{tabular}{@{}l ccc@{}}
\toprule
\textbf{Method} 
& \textbf{Params (B)} 
& \textbf{TFLOPs} 
& \textbf{Runtime (s)} \\
\midrule
UniTime-7B~\cite{li2025universal} & 7.61 & 579 & 37.6 \\
 \textbf{UniversalVTG} & \textbf{0.06} & \textbf{0.0865} & \textbf{0.0886} \\
\bottomrule
\end{tabular}%
}
\end{table}
\footnotetext{Query unification is a one-time \emph{offline} step excluded from VTG inference. Using Qwen3-4B~\cite{yang2025qwen3} on the same A40, unifying a query takes 0.34\,s (0.136 TFLOPs). See Supp.\ for efficiency trade-offs.}

\subsection{Ablation Studies}
\label{sec:ablation_unifier}
We conduct a series of ablation experiments to isolate the contributions of our \textbf{Query Unifier} and \textbf{foundation-style pretraining}. Results are summarized in Table~\ref{tab:ablation_unifier}.

\noindent\textbf{Query Unification vs. Query Quality.} 
A critical question is whether our gains stem from ``better'' text queries or from the model learning a more robust cross-modal alignment. In the second row of Table~\ref{tab:ablation_unifier}, we take a model trained via \emph{naive joint training} and evaluate it using queries converted by the Unifier. We observe a significant performance drop. 
This suggests that simply improving query phrasing at test time is insufficient; rather, the model must be trained on the canonicalized semantic space to resolve the negative transfer inherent in heterogeneous benchmarks and instead unlock generalization through shared grounding logic.

\begin{table}[H]
\centering
\caption{Ablation study of the Query Unifier and Large-Scale Pretraining. We evaluate the impact of mapping queries to a \textbf{unified semantic space} during training (T) and inference (I). All values are reported as Avg.\ R@1.}
\label{tab:ablation_unifier}
\resizebox{\columnwidth}{!}{
\setlength{\tabcolsep}{5.5pt}
\begin{tabular}{ccc ccccc}
\toprule
\multicolumn{2}{c}{\textsc{Unifier}} & \multirow{2}{*}{\makecell{\textsc{Pre-}\\ \textsc{train}}} & \multirow{2}{*}{\makecell{\textsc{GoalStep-}\\ \textsc{StepGrounding}}} & \multirow{2}{*}{\makecell{\textsc{Ego4D-}\\ \textsc{NLQ}}} & \multirow{2}{*}{\textsc{TACoS}} & \multirow{2}{*}{\makecell{\textsc{Charades-}\\ \textsc{STA}}} & \multirow{2}{*}{\makecell{\textsc{ActivityNet-}\\ \textsc{Captions}}} \\
\cmidrule(r){1-2}
\textsc{T} & \textsc{I} & & & & & & \\
\midrule
$\times$ & $\times$ & $\times$ & 22.22 & 20.16 & 55.16 & 61.68 & 38.11 \\ 
$\times$ & \checkmark & $\times$ & 20.39 & 17.21 & 54.54 & 59.91 & 35.57 \\ 
\checkmark & \checkmark & $\times$ & 23.10 & 21.14 & 58.76 & 63.25 & \textbf{40.14} \\ 
\rowcolor{gray!10} \checkmark & \checkmark & \checkmark & \textbf{24.44} & \textbf{24.81} & \textbf{58.91} & \textbf{65.15} & 39.39 \\ 
\bottomrule
\end{tabular}
}
\vspace{-3.5mm}
\end{table}

\noindent\textbf{Full Unification and Pretraining Scalability.} 
The third row demonstrates that applying the Unifier at \emph{both training and inference} provides a substantial leap over the baseline. This confirms that standardizing the instruction space successfully mitigates stylistic friction. Finally, incorporating our harmonized \textbf{1M+ query--segment corpus} for foundation-style pretraining (last row) yields the strongest overall results. This highlights that the ``value-add'' of our structural harmonization is the scale it unlocks for lightweight architectures.

\noindent\textbf{Efficiency of the Query Unifier.}
We emphasize that the Query Unifier is a computationally inexpensive component. Its primary function is the simple conversion of short text fragments (e.g., mapping ``person doing X'' or ``Did I do X?'' to a single canonical style). While this ensures linguistic consistency, it incurs negligible overhead compared to the visual backbone. We provide a thorough analysis of varying unifier models and their efficiency trade-offs in the Supplementary Material.

\section{Conclusion}
We presented \textsc{UniversalVTG}, a single lightweight foundation model for video temporal grounding that remains reliable under substantial cross-dataset heterogeneity. Motivated by practical deployment on long videos, we first establish an efficient shared visual encoder as a common interface across benchmarks. However, we find that naïve joint training across datasets still yields negative transfer: even with a shared architecture, mismatched query styles and annotation conventions hinder optimization. We address this with offline \emph{semantic canonicalization} via a Query Unifier, which standardizes language supervision and enables effective cross-dataset pretraining with a single shared checkpoint.

Our results suggest that scaling supervision and reducing cross-dataset linguistic friction can be more effective than scaling model size for VTG. Looking ahead, we plan to extend the unified training interface to broader temporal video understanding tasks (e.g., temporal action localization and action segmentation), explore end-to-end training that jointly adapts the visual and text encoders alongside the grounding head, and investigate dataset curation and pseudo-labeling pipelines to harvest diverse query--segment pairs from unstructured video for larger-scale pretraining. To support reproducibility and future research, we will release our code, trained checkpoints, and training pipeline.


%
%
\bibliographystyle{splncs04}
\bibliography{main}

\clearpage
\appendix
\setcounter{section}{0}
\setcounter{subsection}{0}
\setcounter{table}{0}
\setcounter{figure}{0}
\setcounter{equation}{0}
\renewcommand{\thesection}{\Alph{section}} 
\renewcommand{\thetable}{S\arabic{table}}  
\renewcommand{\thefigure}{S\arabic{figure}} 
\renewcommand{\theequation}{S\arabic{equation}} 
\renewcommand{\theHsection}{supp.\Alph{section}}
\renewcommand{\theHtable}{supp.\arabic{table}}
\renewcommand{\theHfigure}{supp.\arabic{figure}}
\renewcommand{\theHequation}{supp.\arabic{equation}}
\markboth{Supplementary Material}{Supplementary Material}

\begin{table}[b!] 
\centering
\caption{Performance and unifier compute metrics across multiple datasets. Consistent with the main manuscript and following established conventions, performance is reported as Average Recall@1 at IoU thresholds of 0.3 and 0.5 for long-form video datasets (GoalStep~\cite{song2023ego4d}, NLQ~\cite{grauman2022ego4d}, TACoS~\cite{regneri2013grounding}), and at 0.5 and 0.7 for short-form datasets (Charades~\cite{sigurdsson2016hollywood}, Activitynet~\cite{krishna2017dense}).}
\label{tab:unifier_compute}
\resizebox{\textwidth}{!}{%
\setlength{\tabcolsep}{1.5pt}
\begin{tabular}{cc ccccc ccc} 
\toprule
\multirow{3}{*}{Method} & \multirow{3}{*}{Unifier} & \multirow{3}{*}{GoalStep} & \multirow{3}{*}{NLQ} & \multirow{3}{*}{TACoS} & \multirow{3}{*}{\begin{tabular}{@{}c@{}}Cha.-\\STA\end{tabular}} & \multirow{3}{*}{\begin{tabular}{@{}c@{}}Anet-\\Cap.\end{tabular}} & \multicolumn{3}{c}{Unifier Compute} \\
\cmidrule(lr){8-10}
& & & & & & & \multicolumn{2}{c}{Runtime (s)} & \multirow{2}{*}{TFLOPs} \\
\cmidrule(lr){8-9}
& & & & & & & Prefill & Decode & \\
\midrule
\multirow{2}{*}{Naive Joint Training} & --            & 22.22 & 20.16 & 55.16 & 61.68 & 38.11 & --    & --    & --    \\
                                      & GPT5          & 20.39 & 17.21 & 54.54 & 59.91 & 35.57 & --    & --    & --    \\
\midrule
\multirow{5}{*}{UniversalVTG}         & $\times$      & 22.69 & 21.05 & 55.55 & 63.03 & 37.95 & --    & --    & --    \\
                                      & GPT5          & 24.44 & 24.81 & 58.91 & 65.15 & 39.39 & --    & --    & --    \\
                                      & Qwen3-4B      & 24.56 & 24.53 & 58.61 & 65.19 & 38.57 & 0.105 & 0.340 & 0.136 \\
                                      & Qwen3-30B     & 24.24 & 24.61 & 58.77 & 64.94 & 39.31 & 0.333 & 1.38  & 0.84  \\
                                      & Llama3.1-70B  & 24.77 & 24.50 & 58.99 & 64.94 & 38.28 & 0.14  & 0.891 & 1.96  \\
\bottomrule
\end{tabular}%
}
\end{table}


\section{Analysis of the Unifier Module}
\label{sec:supp_unifier}

As introduced in Section 4.2 of the main paper, the Unifier module can be instantiated with any Large Language Model (LLM). Its primary role is to convert diverse incoming queries into a standardized, canonical form to facilitate universal video temporal grounding. The exact prompt utilized to guide the LLM for this conversion is provided in Prompt~\ref{prompt:unifier}. 

During the two-stage training pipeline (detailed in Section 4.4), we employed different LLMs to process the data. For the pretraining phase (Stage-I), Llama3.1-70B~\cite{grattafiori2024llama} was used as the Unifier to canonicalize the pretraining text data. Subsequently, for the fine-tuning phase (Stage-II), we utilized GPT-5~\cite{singh2025openai} (\texttt{gpt-5.2-2025-12-11}). 

\noindent\textbf{Unifier Integration, Agnosticism, and Efficiency.} To fully contextualize the Unifier's impact, Table~\ref{tab:unifier_compute} presents a detailed breakdown of different training paradigms and Unifier configurations. As shown in the first two rows, simply attaching a Unifier during inference to a naively joint-trained model does not yield meaningful improvements, corroborating our findings in Section 5.4 and Table 8 of the main paper. Conversely, our UniversalVTG framework inherently benefits from being trained on a canonical text space alongside massive data. Even in a sub-optimal inference setup where the Unifier is entirely omitted (denoted as `$\times$' in Table~\ref{tab:unifier_compute}), UniversalVTG consistently outperforms the naive joint training baseline. 

However, the framework's full potential is unlocked when the Unifier is reintroduced at inference time. To thoroughly investigate the flexibility and potential computational overhead of this integration, we evaluated UniversalVTG by swapping the Unifier at inference time with a variety of models, scaling from small (Qwen3-4B~\cite{yang2025qwen3}) to medium (Qwen3-30B~\cite{yang2025qwen3}), and large (Llama3.1-70B~\cite{grattafiori2024llama}, GPT-5~\cite{singh2025openai}) architectures. Crucially, the core UniversalVTG checkpoint remains completely untouched during these evaluations. We utilize the exact same model weights obtained after Stage-I (pretrained with the Llama3.1-70B~\cite{grattafiori2024llama} unifier) and Stage-II (fine-tuned with the GPT-5~\cite{singh2025openai} unifier), demonstrating that the Unifier can be strictly treated as a plug-and-play module at inference without any need for downstream retraining.

First, the downstream grounding performance remains remarkably consistent across all tested models, demonstrating that UniversalVTG is fundamentally agnostic to the specific choice of Unifier. Even the highly lightweight 4B model achieves results strictly competitive with massive proprietary models. This robust performance is expected: converting text to a canonicalized format is a straightforward task, and modern LLMs can easily follow our explicit instruction prompt to execute it. 

Second, our profiling demonstrates that the added computational burden is minimal. Using the same hardware setup employed for our grounding-head efficiency comparison (a single A40 GPU, as in Table 7), the text conversion process requires only $\sim$1 second even when using the largest model we tested (Llama3.1-70B), and less than half a second when using the lightweight 4B model. Crucially, as shown in Table~\ref{tab:efficiency_comparison}, this overhead represents a negligible fraction of the total inference compute when compared to the heavy computational cost of video feature extraction.

Ultimately, this study confirms that alongside our forthcoming publicly released checkpoints, researchers and practitioners can flexibly deploy an LLM that best fits their specific needs and computational constraints as the Unifier, without sacrificing grounding accuracy or incurring prohibitive computational costs.

\section{Efficiency Comparison with State-of-the-Art MLLMs}
\label{sec:supp_efficiency}

As demonstrated in Table 6 of the main manuscript, UniversalVTG achieves temporal grounding performance comparable to UniTime-7B~\cite{li2025universal}, a state-of-the-art Multimodal Large Language Model (MLLM) specifically trained for video temporal grounding. Due to the strict page limits of the main manuscript, we were unable to include the full, granular breakdown of our efficiency profiling. Building upon the preliminary summary metrics introduced in Table 7 of the main paper, Table~\ref{tab:efficiency_comparison} provides this complete, original profiling of computational costs. This detailed breakdown, conducted concurrently with our primary evaluations, isolates the parameter count, TFLOPs, and inference runtime, explicitly decoupling the visual encoder (feature extraction) from the grounding module.

\begin{table}[t] 
\centering
\caption{Efficiency comparison of parameter count, TFLOPs, and runtime across different video lengths. Vis. Enc.: Visual Encoder; Feat. Ext.: Feature Extraction.}
\label{tab:efficiency_comparison}
\resizebox{\textwidth}{!}{%
\setlength{\tabcolsep}{3pt}
\begin{tabular}{l cc ccc ccc}
\toprule
\multirow{2}{*}{Method} & \multicolumn{2}{c}{Parameter Count (B)  \textbf{($\downarrow$)}} & \multicolumn{3}{c}{TFLOPs \textbf{($\downarrow$)}} & \multicolumn{3}{c}{Runtime (s)  \textbf{($\downarrow$)}} \\
\cmidrule(lr){2-3} \cmidrule(lr){4-6} \cmidrule(lr){7-9}
& Vis. Enc. & Grounder & Feat. Ext. & Grounding & Total & Feat. Ext. & Grounding & Total \\
\midrule
\multicolumn{9}{c}{\textit{8.3-Minute Video}} \\
\midrule
UniTime~\cite{li2025universal}      & 0.68 & 7.61 & 2184 & 579    & 2763 & 72.1  & 35.4   & 107.6 \\
\rowcolor{gray!15} UniversalVTG & 0.32 & 0.06 & 175  & 0.0865 & 175  & 11.1  & 0.0886 & 11.2  \\
\midrule
\multicolumn{9}{c}{\textit{15.0-Minute Video}} \\
\midrule
UniTime~\cite{li2025universal}      & 0.68 & 7.61 & 5384 & 588    & 5972 & 154.6 & 80.2   & 234.8 \\
\rowcolor{gray!15} UniversalVTG & 0.32 & 0.06 & 317  & 0.155  & 317  & 19.3  & 0.0928 & 19.4  \\
\bottomrule
\end{tabular}%
}
\end{table}

To ensure a rigorous and strictly fair comparison, both models were evaluated on the exact same videos at two distinct durations: an 8.3-minute (500 seconds) video, which mirrors the average video length in the Ego4D-NLQ dataset, and a longer 15-minute video, representing a shorter sample from the GoalStep dataset. Runtime and compute measurements were extracted by strictly adhering to UniTime's official evaluation codebase. For both models, the feature extraction phase processed frames in batches of 16. Specifically, UniTime employs a fixed total pixel budget that dynamically balances spatial resolution and temporal sampling rate per video. While its nominal target is 2 frames per second (fps), the effective rate varies with video duration. In contrast, UniversalVTG consistently processes the temporal sequence at a fixed 2 fps using a lightweight Perception Encoder~\cite{bolya2025perception}, deliberately avoiding heavy video backbones. By feeding the identical video into both pipelines and measuring exactly how each framework natively samples and extracts frames, the metrics presented in Table~\ref{tab:efficiency_comparison} directly reflect the true real-world computational footprint of each method.

The profiling results unequivocally demonstrate the computational superiority of UniversalVTG. Across all measured metrics---parameter count, FLOPs, and runtime---our framework is vastly lighter and faster. While video feature extraction naturally remains the dominant computational bottleneck for both methods, UniversalVTG utilizes an exceptionally lightweight VTG-specific grounding head. As shown in Table~\ref{tab:efficiency_comparison}, the grounding process executes in mere fractions of a second (e.g., $\sim$0.09 seconds), even on the 15-minute video. 

This architectural efficiency yields a profound advantage in practical, real-world deployments. In typical multi-query scenarios, the heavy feature extraction process only needs to be executed once per video. Any subsequent text queries rely solely on the grounding head. Because UniversalVTG's grounding operates in millisecond units, the marginal cost of additional queries is effectively negligible. Consequently, UniversalVTG serves as a highly attractive and scalable alternative to massive MLLMs, delivering comparable state-of-the-art accuracy at a fraction of the computational cost.

\begin{table}[t]
\centering

\begin{minipage}[c]{0.45\textwidth}
\centering
\caption{Cross-dataset evaluation (AVG R@1). Rows indicate the \textbf{training} dataset and columns indicate the \textbf{evaluation} dataset. }
\label{tab:cross_dataset}
\resizebox{\linewidth}{!}{%
\begin{tabular}{@{}l cccc@{}}
\toprule
\makecell[l]{\textbf{Test $\rightarrow$} \\ \textbf{Train $\downarrow$}} 
& \makecell{\textbf{NLQ}} 
& \makecell{\textbf{TACoS}} 
& \makecell{\textbf{Charades} \\ \textbf{-STA}} 
& \makecell{\textbf{ANet} \\ \textbf{-Cap.}} \\
\midrule
NLQ          & \textbf{19.55} & 20.18 & 23.50 & 13.16 \\
TACoS        & 3.39 & \textbf{56.70} & 13.23 & 10.78 \\
Cha.-STA     & 4.35 & 12.92 & \textbf{62.35} & 9.35 \\
ANet-Cap.    & 3.93 & 13.97 & 22.43 & \textbf{39.52} \\
\midrule
\midrule
Prior Unified SOTA~\cite{lin2023univtg}    & 9.64 & 49.78 & 49.37 & 31.98 \\
\rowcolor{gray!15} UniversalVTG    & 24.81 & 58.91 & 65.15 & 39.39 \\
\bottomrule
\end{tabular}%
}
\end{minipage}\hfill
\begin{minipage}[c]{0.52\textwidth}
\centering
\includegraphics[width=\linewidth]{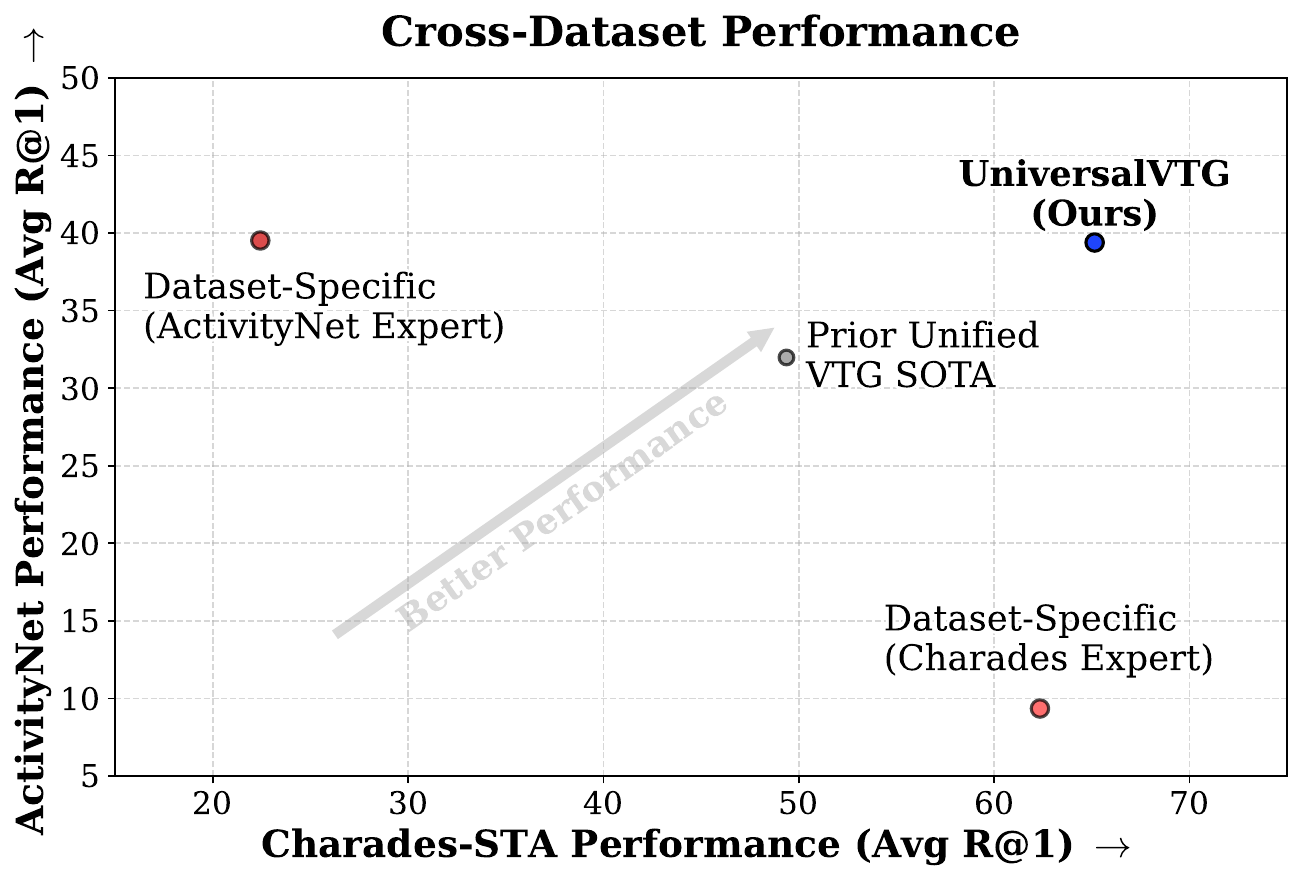}
\captionof{figure}{Additional Cross-Dataset Performance.}
\label{fig:crossdataset}
\end{minipage}
\end{table}

\section{Complete Cross-Dataset Evaluation and Domain Transfer}
\label{sec:supp_cross_dataset}

Table~\ref{tab:cross_dataset} provides the complete cross-dataset evaluation matrix, which was summarized in Table 2 of the main manuscript strictly due to strict page limitations. While the top four rows reproduce those primary findings, we provide essential methodological details here regarding our original experimental setup that were omitted for brevity. Specifically, every model in this evaluation---including the dataset-specific experts---shares the exact same unified visual representation space using the PerceptionEncoder~\cite{bolya2025perception}. Crucially, establishing this unified visual feature space is what fundamentally enabled our direct cross-dataset validation in the first place. By ensuring the underlying visual representations remained strictly consistent across diverse domains during these evaluations, we isolated the true impact of dataset-specific training versus unified training without the confounding variable of mismatched visual backbones.

In this experimental framework, we trained dataset-specific expert models and evaluated each on unseen domains, where rows indicate the training dataset (the expert's source domain) and columns represent the evaluation datasets. The results clearly demonstrate that despite sharing a unified visual representation space, dataset-specific models transfer remarkably poorly to unseen domains. This domain gap is most pronounced when models trained exclusively on exocentric video datasets (TACoS~\cite{regneri2013grounding}, Charades-STA~\cite{sigurdsson2016hollywood}, and ActivityNet-Captions~\cite{krishna2017dense}) are evaluated on the egocentric Ego4D-NLQ~\cite{grauman2022ego4d} dataset. In these cases, performance drops precipitously, highlighting the severe limitations of specialized models when confronted with diverse, real-world temporal grounding scenarios.

To further visualize this phenomenon across all temporal scales, Figure~\ref{fig:crossdataset} plots the cross-dataset performance comparison specifically between ActivityNet-Captions and Charades-STA. This directly complements Figure 3 of the main manuscript; whereas Figure 3 illustrated the severe domain transfer bottleneck across our \textit{long-form} video pair (TACoS and Ego4D-NLQ), Figure~\ref{fig:crossdataset} provides the corresponding, previously omitted analysis for our \textit{short-form} video datasets. As depicted in the scatter plot, the dataset-specific experts remained heavily biased toward their training domains, suffering massive performance drops when evaluated on the alternate dataset. 

To benchmark our framework against these limitations, we compared our single, unified UniversalVTG model against both the dataset-specific experts and the prior state-of-the-art unified VTG model~\cite{lin2023univtg}. As shown in the final rows of Table~\ref{tab:cross_dataset} and visually confirmed in Figure~\ref{fig:crossdataset}, UniversalVTG substantially outperformed the prior unified SOTA across all evaluated benchmarks. More crucially, despite being a single foundational model handling all domains simultaneously, UniversalVTG surpassed the grounding accuracy of nearly all dataset-specific experts on their respective home datasets. Ultimately, this complete evaluation confirms that UniversalVTG effectively bridged massive domain gaps across both short and long video formats without requiring dataset-specific architectures or isolated training pipelines.

\begin{table}[h] 
\centering
\caption{Comparison across five VTG benchmarks for Recall@5 (R@5). R@5 is reported at multiple IoU thresholds with the average shown in each block. ``--'' indicates results not available or not applicable. Models marked with $\blacksquare$ utilize shared weights across all benchmarks (a single foundational model), whereas $\square$ denotes models requiring separate weights fine-tuned per dataset.}
\label{tab:supp_r5_results}
\resizebox{\textwidth}{!}{%
\begin{tabular}{@{}lc ccc ccc ccc ccc ccc@{}}
\toprule
\multirow{3}{*}{\textbf{Method}} & \multirow{3}{*}{\makecell{\textbf{Shared} \\ \textbf{Weights}}}
& \multicolumn{9}{c}{\textbf{Long-Form Videos}}
& \multicolumn{6}{c}{\textbf{Short-Form Videos}} \\
\cmidrule(lr){3-11} \cmidrule(lr){12-17}
& & \multicolumn{3}{c}{\textbf{GoalStep}}
& \multicolumn{3}{c}{\textbf{Ego4D-NLQ}}
& \multicolumn{3}{c}{\textbf{TACoS}}
& \multicolumn{3}{c}{\textbf{Cha.-STA}}
& \multicolumn{3}{c}{\textbf{ANet-Cap.}} \\
\cmidrule(lr){3-5} \cmidrule(lr){6-8} \cmidrule(lr){9-11} \cmidrule(lr){12-14} \cmidrule(lr){15-17}
& & 0.3 & 0.5 & Avg.
  & 0.3 & 0.5 & Avg.
  & 0.3 & 0.5 & Avg.
  & 0.5 & 0.7 & Avg.
  & 0.5 & 0.7 & Avg. \\
\midrule

2D-TAN~\cite{zhang2020learning}      & $\square$ & -- & -- & -- & 12.89 & 5.88 & 9.39 & 69.11 & 57.31 & 63.21 & 89.14 & 61.13 & 75.14 & 78.80 & 60.85 & 69.83 \\
M-DETR~\cite{lei2021detecting}       & $\square$ & -- & -- & -- & 23.23 & 13.37 & 18.30 & -- & -- & -- & -- & -- & -- & -- & -- & -- \\
CONE~\cite{hou2022cone}              & $\square$ & -- & -- & -- & 30.33 & 18.02 & 24.18 & -- & -- & -- & -- & -- & -- & -- & -- & -- \\
UnLoc-L~\cite{yan2023unloc}          & $\square$ & -- & -- & -- & -- & -- & -- & -- & -- & -- & 88.20 & 61.10 & 74.65 & 79.20 & 61.30 & 70.25 \\
RGNet~\cite{hannan2024rgnet}                                & $\square$ & -- & -- & -- & 34.02 & 22.89 & 28.46 & -- & -- & -- & -- & -- & -- & -- & -- & -- \\
SnAG~\cite{snag}                     & $\square$ & - & - & - & 38.39 & 27.44 & 32.92 & 81.15 & 70.66 & 75.91 & 92.55 & 64.11 & 78.33 & 81.71 & 63.41 & 72.56 \\
DeCafNet~\cite{lu2025decafnet}       & $\square$ & 47.27 & 40.40 & 43.84 & 38.85 & 28.27 & 33.56 & 81.05 & 71.13 & 76.14 & 91.53 & 72.96 & 82.25 & -- & -- & -- \\
HieraMamba~\cite{an2025hieramamba}  & $\square$ & -- & -- & -- & 40.82 & 29.96 & 35.39 & 83.75 & 74.28 & 79.02 & -- & -- & -- & -- & -- & -- \\

\midrule
\rowcolor{gray!15} \textbf{UniversalVTG} & $\blacksquare$
& \textbf{54.35} & \textbf{47.22} & \textbf{50.79}
& \textbf{53.47} & \textbf{41.52} & \textbf{47.50}
& \textbf{85.88} & \textbf{78.66} & \textbf{82.27}
& \textbf{93.66} & \textbf{77.98} & \textbf{85.82}
& \textbf{84.73} & \textbf{63.64} & \textbf{74.19} \\

\bottomrule
\end{tabular}%
}
\end{table}

\section{Comprehensive Recall@5 Performance Metrics}
\label{sec:supp_recall5}

In Table 6 of the main manuscript, we established the strong temporal grounding capabilities of UniversalVTG primarily using the strict Recall@1 (R@1) metric due to page limits. However, because many prior dedicated video temporal grounding architectures standardly report Recall@5 (R@5) across various Intersection over Union (IoU) thresholds, we recorded these metrics concurrently during our original evaluation phase. We provide the complete R@5 results here to ensure a fully comprehensive comparison against the literature.

Table~\ref{tab:supp_r5_results} presents the full R@5 performance for UniversalVTG alongside prior baselines for which these specific metrics were publicly available. Serving as a direct complement to the main paper's results, these metrics further validate our primary findings. Consistent with our R@1 analysis, UniversalVTG demonstrated dominant grounding accuracy across both long-form and short-form video datasets. Crucially, it achieved state-of-the-art R@5 performance across the board, significantly outperforming highly specialized, dataset-specific expert models despite operating entirely on a single, unified set of weights.

\section{Qualitative Analysis}
\label{sec:supp_qualitative}

To further illustrate the robust generalization capabilities of our framework, Figure~\ref{fig:qualitative} presents qualitative temporal grounding results across the evaluated datasets. These examples highlight UniversalVTG's capacity to seamlessly operate on highly diverse video domains and text queries without requiring domain-specific fine-tuning.

As shown in the figure, the visual data spans a wide spectrum of characteristics. The examples encompass both egocentric views (GoalStep, Ego4D-NLQ) and exocentric perspectives (TACoS, Charades-STA, ActivityNet-Captions). Furthermore, the model effectively handles extreme variations in video duration, successfully processing long-form untrimmed videos lasting over ten minutes (e.g., 752 seconds for GoalStep and 651 seconds for TACoS) alongside short-form clips lasting less than 20 seconds (e.g., 17 seconds for ActivityNet-Captions).

Equally important is the framework's adaptability to diverse linguistic queries. The grounding pipeline gracefully manages a variety of semantic structures, including concise imperative commands (``peel potatoes''), first-person interrogatives (``What did I put in the chopping machine?''), multi-object interactions (``Take out a pomegranate and a cutting board''), casual third-person action descriptions (``person turn a light on''), and highly detailed, attribute-rich sentences (``A young lady is gripping a black and silver punching bag between her legs.'').

Despite these massive shifts in the visual domain, temporal scale, and linguistic structure, UniversalVTG accurately localizes the relevant target moments with high precision, successfully mapping the correct temporal boundaries using just a single foundational checkpoint.

\begin{figure}[t]
    \centering
    \begin{minipage}[t]{0.495\textwidth}
        \centering
        \includegraphics[width=\linewidth]{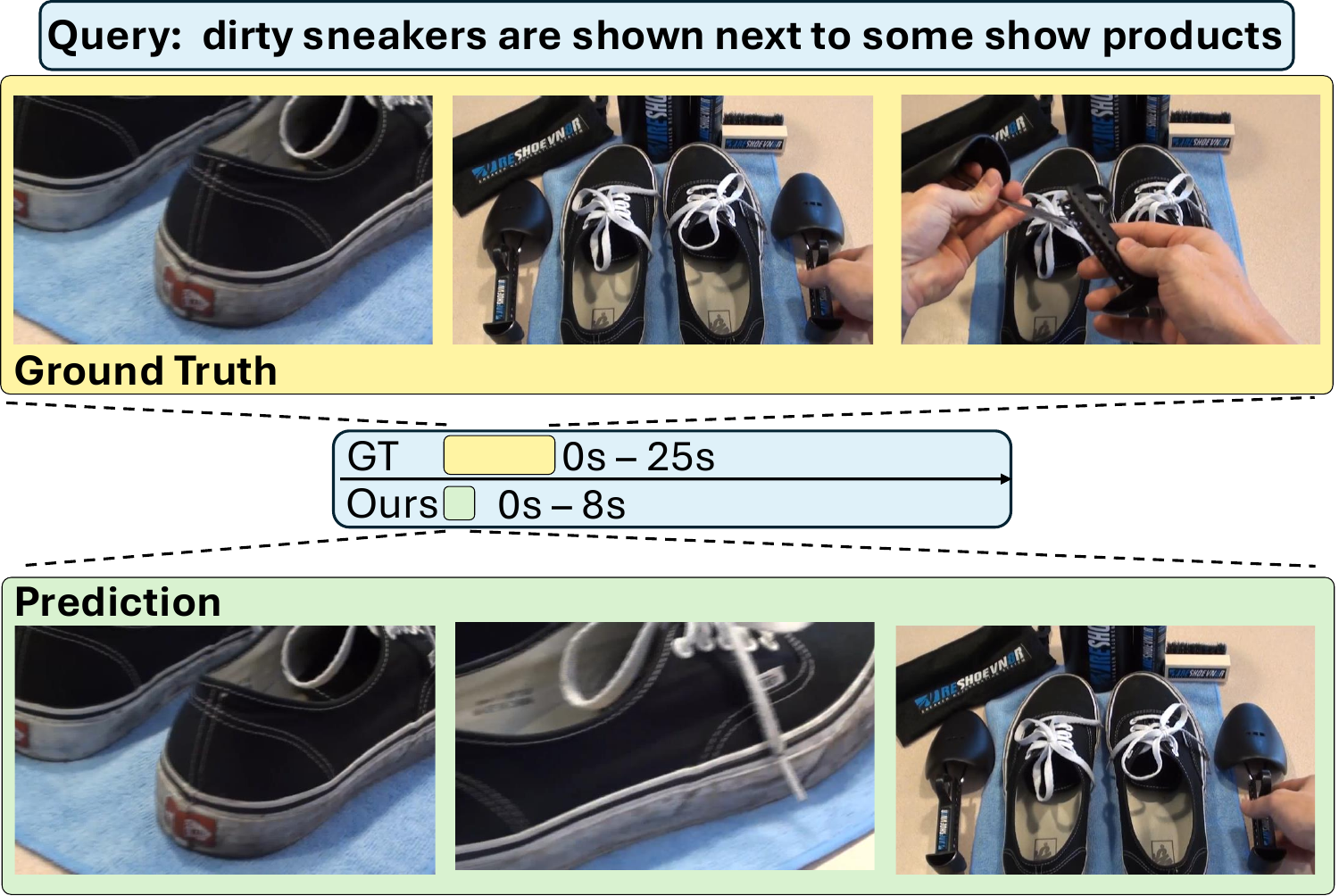}
    \end{minipage}\hfill
    \begin{minipage}[t]{0.495\textwidth}
        \centering
        \includegraphics[width=\linewidth]{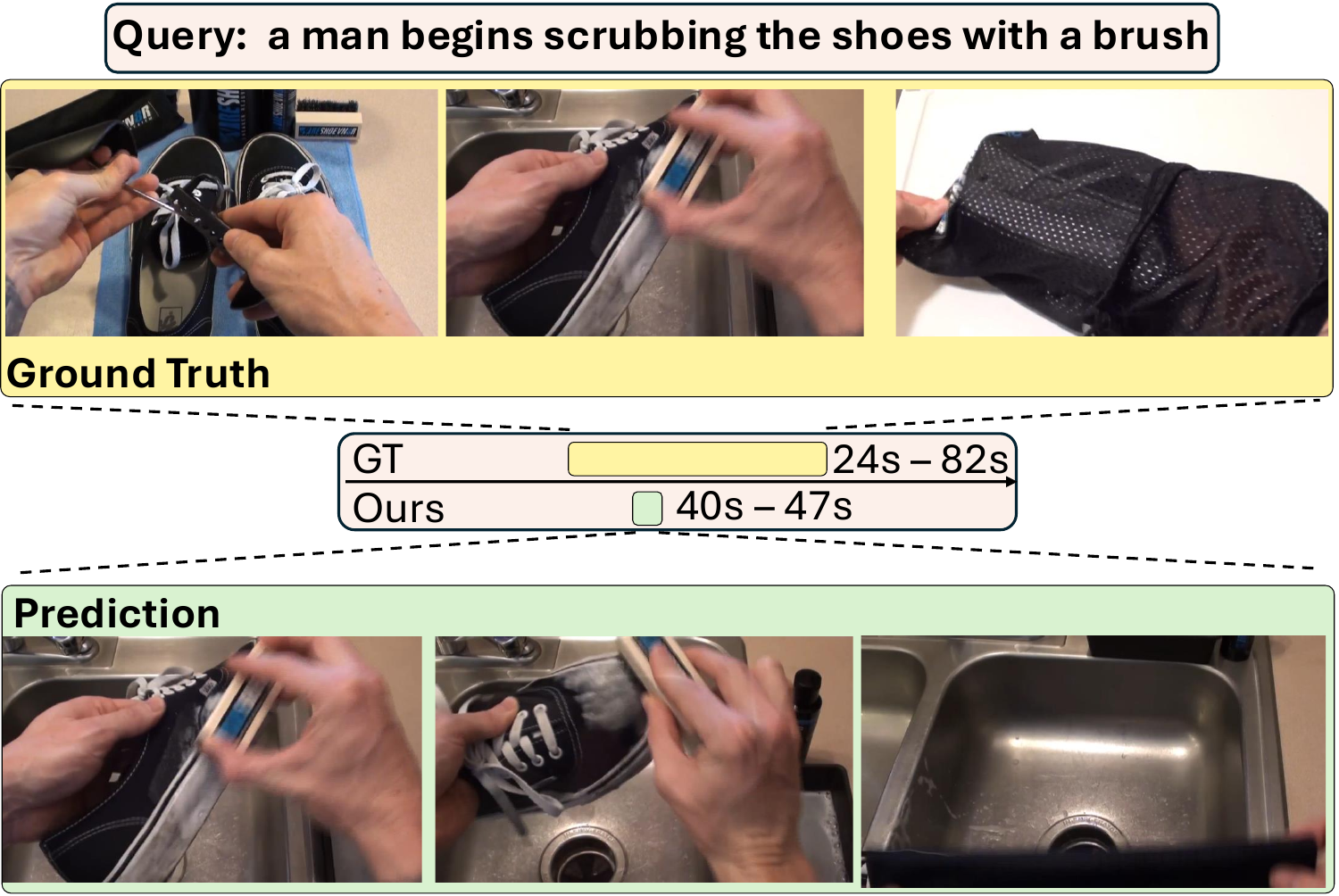}
    \end{minipage}
    
    \caption{Qualitative results of failure cases.}
    \label{fig:failures}
\end{figure}

\section{Failure Case Analysis}
\label{sec:supp_failures}

While UniversalVTG demonstrates strong generalization across diverse domains, an analysis of its failure cases reveals areas where the model faces certain challenges. Specifically, the framework occasionally finds it difficult to associate the continuity of an object's state or an action's broader context across camera cuts, particularly when facing abrupt viewpoint shifts and camera transitions. Figure~\ref{fig:failures} illustrates two typical examples of this behavior.

In the first failure case (Figure~\ref{fig:failures}, left), the query asks to localize when "dirty sneakers are shown next to some shoe products." UniversalVTG accurately detects the start of the event (0 seconds) when the dirty shoes are clearly visible from a side-view angle. However, at the 8-second mark, the video cuts to a top-down view. While a human observer naturally infers that these are the exact same shoes continuing to be displayed, the top-down angle obscures the dirtiness. Without an explicit mechanism to correlate the object's identity across the scene change, the model terminates its prediction early (predicting 0--8s). Conversely, the ground truth spans from 0 to 25 seconds, capturing the entire continuous preparation sequence until the person visibly moves on to a different task.

A similar pattern of temporal fragmentation is observed in the second case (Figure~\ref{fig:failures}, right), where the query is "a man begins scrubbing the shoes with a brush." The ground-truth annotation encompasses the entire logical sequence of the overarching event: preparing the brush, the physical scrubbing, the subsequent cleaning, and finally placing the shoes into a bag. However, UniversalVTG tends to focus closely on the literal, immediate action. It successfully predicts the exact, narrow window where the brush physically contacts the shoe, but stops predicting when the camera angle shifts or the literal brushing pauses. In this instance, the model does not fully capture the pre- and post-processes associated with the core action.

Ultimately, these examples highlight that while UniversalVTG is highly adept at localizing explicit, visually persistent actions, it can sometimes experience temporal fragmentation when events span across camera cuts or involve complex, multi-step procedures that require broader cognitive reasoning beyond immediate, frame-level visual evidence.

\begin{figure}[h]
    \centering
    \includegraphics[width=0.97\linewidth]{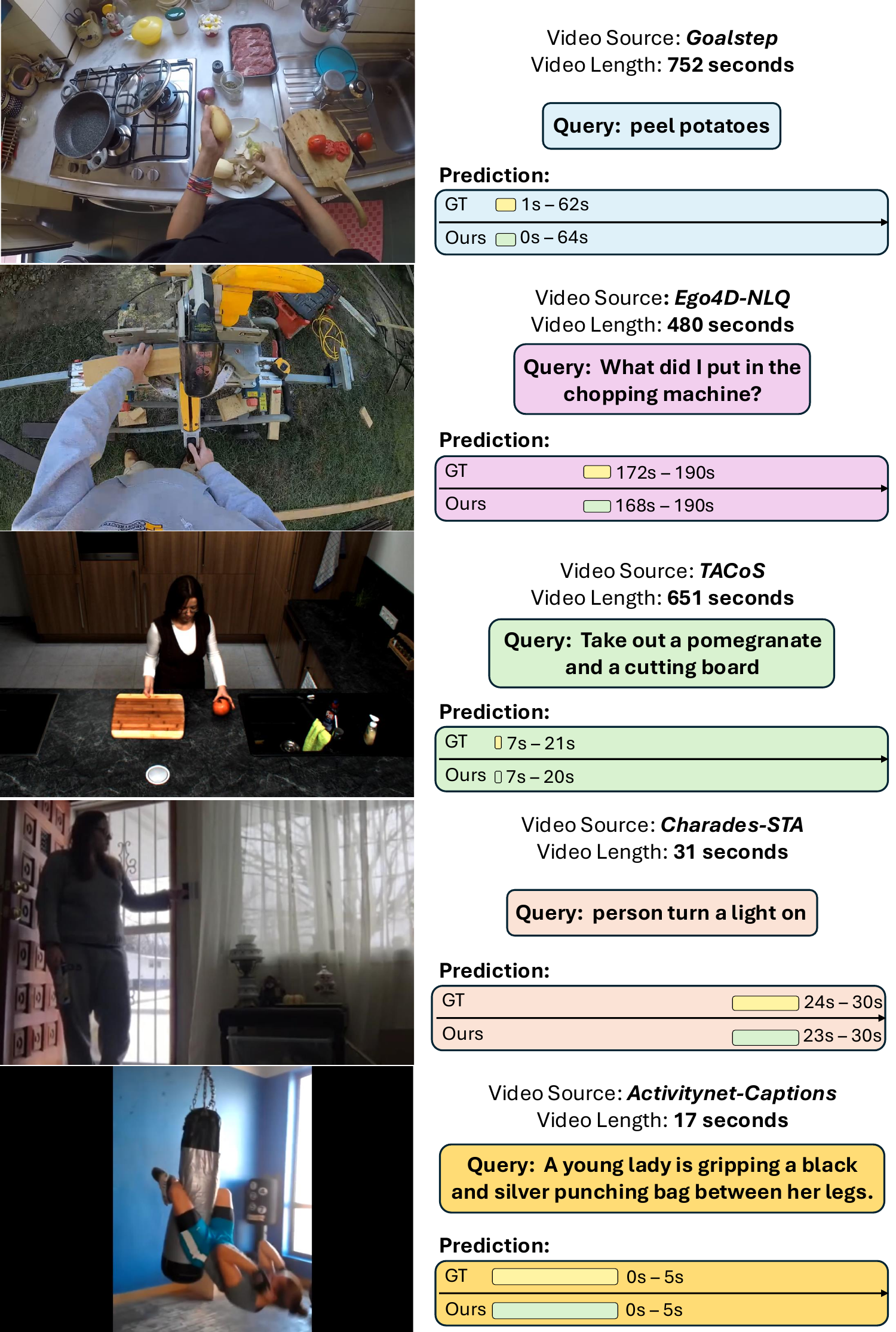}
    \caption{Qualitative grounding results across five diverse VTG benchmarks. UniversalVTG accurately localizes temporal segments across varying camera perspectives, video durations, and linguistic query styles using a single unified model. Ground-truth (GT) and predicted segments (Ours) are denoted for each video.}
    \label{fig:qualitative}
\end{figure}


\begin{promptbox}[label={prompt:unifier}]{System Prompt for Unifier}
\small
\textbf{Unified Video Query Conversion: Preserve All Semantic Meaning}

\vspace{0.5em}
You are given a sentence describing an event in a video from one of the different datasets. Your task is to convert it into a unified format that:
\begin{enumerate}
    \item Preserves ALL semantic meaning, visual cues, objects, attributes, and agent information.
    \item Standardizes the style to a consistent format across all datasets.
    \item Maintains the original perspective (first-person if original is first-person, third-person if original is third-person).
\end{enumerate}

\vspace{0.5em}
\hrule
\vspace{0.5em}

\noindent\textbf{Unified Style Target}
\begin{enumerate}
    \item \textbf{Format (Descriptive statement):} All queries should be phrased as complete, descriptive statements. Directly describe the event to locate (Subject-Verb-Object-Location), which is more natural for temporal matching than questions.
    \item \textbf{Tense (Past tense):} Use past tense verbs (e.g., "was", "put", "took"). This indicates events that have already occurred and can be located temporally.
    \item \textbf{Perspective Preservation:} 
    \begin{itemize}
        \item If original is first-person ("I", "my"): \textbf{Keep first-person}.
        \item If original is third-person ("he", "the woman"): \textbf{Keep third-person}.
        \item \textbf{Do NOT convert} between perspectives. Preserve agent identity explicitly.
    \end{itemize}
    \item \textbf{Completeness:} Use complete, grammatically correct sentences. Capitalize the first letter and end with a period.
    \item \textbf{Visual Grounding:} All information must be purely visually deducible. Do NOT require reasoning about intentions, motivations, or external knowledge.
\end{enumerate}

\vspace{0.5em}
\hrule
\vspace{0.5em}

\noindent\textbf{Critical Preservation Rules}
\begin{itemize}
    \item \textbf{1. Agent/Subject:} Preserve the original agent noun phrase exactly (e.g., "I" $\rightarrow$ "I", "a woman" $\rightarrow$ "a woman"). Agent identity is a strong visual cue.
    \item \textbf{2. Action/Verb:} Preserve the main verb phrase. Do NOT replace specific actions (e.g., "wax down a ski") with generic verbs like "doing" or "acting".
    \item \textbf{3. Objects \& 4. Attributes:} Preserve all key objects (e.g., tiles, ski, ball) and attributes (e.g., "young", "yellow").
    \item \textbf{5. Location/Spatial:} Preserve all location references (e.g., "inside a gym", "on the shelf").
    \item \textbf{6. Temporal Handling:} Avoid ordinal references (first, second) unless converted to visually grounded sequences (e.g., "second outfit" $\rightarrow$ "after changing into a different outfit").
    \item \textbf{7. Question-to-Statement:} Focus on describing the EVENT, not the answer. Use specific event-focused verbs.
\end{itemize}

\vspace{0.5em}
\hrule
\vspace{0.5em}

\noindent\textbf{Statement Structure \& Disallowed Content}
\begin{itemize}
    \item \textbf{Structure:} Subject (Agent) + Verb (Action) + Object + Modifiers (Location, Manner).
    \item \textbf{Do NOT include:} Reasoning about intentions ("why"), external knowledge (brand names), off-screen information, future predictions, or vague placeholders ("somewhere", "something").
\end{itemize}

\vspace{0.5em}
\hrule
\vspace{0.5em}

\noindent\textbf{Conversion Pipeline Guidelines}
\begin{enumerate}
    \item \textbf{Identify elements:} Agent, Action, Objects, Location, Attributes, Temporal relations.
    \item \textbf{Preserve perspective:} First-person stays first-person; third-person stays third-person.
    \item \textbf{Convert to statement:} Use specific event verbs. Avoid weak verbs ("was", "had") unless part of a compound event.
    \item \textbf{Convert to past tense:} "is" $\rightarrow$ "was", "takes" $\rightarrow$ "took".
    \item \textbf{Output Requirements:} Provide ONLY the converted statement. No explanations, no original sentences, no commentary. A single, complete, grammatically correct sentence ending in a period.
\end{enumerate}

\end{promptbox}


\clearpage
\end{document}